\documentclass{article}

\usepackage{arxiv}

\usepackage[utf8]{inputenc} 
\usepackage[T1]{fontenc}    
\usepackage{hyperref}       
\usepackage{url}            
\usepackage{booktabs}       
\usepackage{nicefrac}       
\usepackage{microtype}      
\usepackage{lipsum}		
\usepackage{graphicx}
\usepackage{doi}

\usepackage{multirow}
\usepackage{diagbox}
\usepackage{placeins}
\usepackage{xcolor}
\usepackage{bm}

\usepackage{soul}
\usepackage{subfig}
\usepackage{graphicx}
\usepackage[cmex10]{amsmath}
\usepackage{amsfonts}
\usepackage{newpxtext}
\usepackage{newpxmath}
\usepackage[ruled,vlined]{algorithm2e}

\title{Modeling Regime Shifts in Multiple Time Series}

\author{Etienne Gael Tajeuna \\
	Department of Computer Science\\
	University of Sherbrooke\\
	Sherbrooke, Quebec, Canada \\
	\texttt{etienne.gael.tajeuna@usherbrooke.ca} \\
	\And
	Mohamed Bouguessa \\
	Department of Computer Science\\
	University of Quebec at Montreal\\
	Montreal, Quebec, Canada \\
	\texttt{bouguessa.mohamed@uqam.ca}  \\
	\And
	Shengrui Wang \\
	Department of Computer Science\\
	University of Sherbrooke\\
	Sherbrooke, Quebec, Canada \\
	\texttt{shengrui.wang@usherbrooke.ca} \\
}

\date{}

\hypersetup{
pdfauthor={Tajeuna et al.},
}

\begin{document}
\maketitle

\begin{abstract}
  We investigate the problem of discovering and modeling regime shifts in an ecosystem comprising multiple time series known as co-evolving time series. Regime shifts refer to the changing behaviors exhibited by series at different time intervals. Learning these changing behaviors is a key step toward time series forecasting. While advances have been made, existing methods suffer from one or more of the following shortcomings: (1) failure to take relationships between time series into consideration for discovering regimes in multiple time series; (2) lack of an effective approach that models time-dependent behaviors exhibited by series; (3) difficulties in handling data discontinuities which may be informative. Most of the existing methods are unable to handle all of these three issues in a unified framework. This, therefore, motivates our effort to devise a principled approach for modeling interactions and time-dependency in co-evolving time series. Specifically, we model an ecosystem of multiple time series by summarizing the heavy ensemble of time series into a lighter and more meaningful structure called a \textit{mapping grid}. By using the mapping grid, our model first learns time series behavioral dependencies through a dynamic network representation, then learns the regime transition mechanism via a full time-dependent Cox regression model. The originality of our approach lies in modeling interactions between time series in regime identification and in modeling time-dependent regime transition probabilities, usually assumed to be static in existing work.
\end{abstract}



\section{Introduction}
\label{article2-sec:introduction}
\begin{figure}[tbp]
\centering
\includegraphics[width=4in]{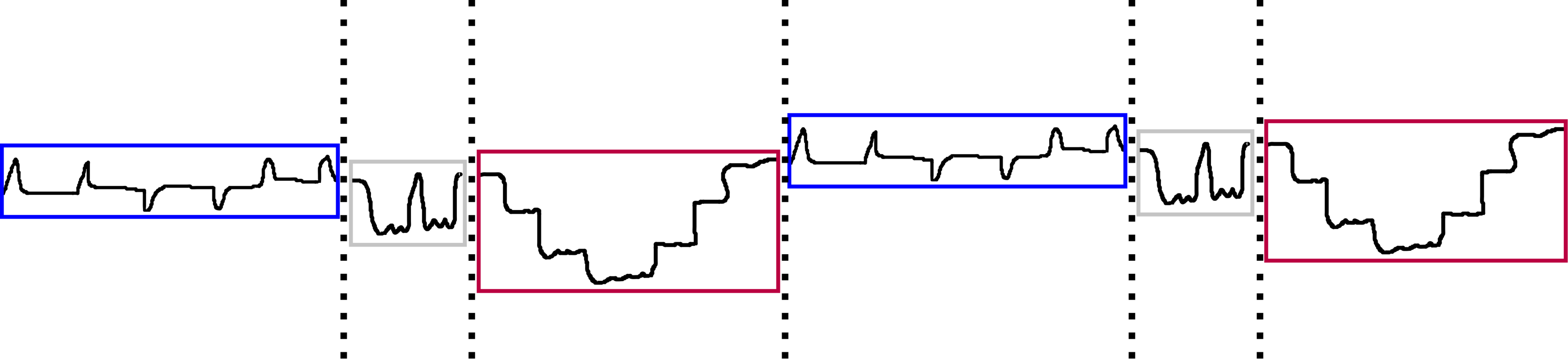}
\caption{ An example of a series with three contiguous regimes. Vertical dashed lines delimit regime time intervals.}
\label{fig:contiguous regimes}
\end{figure}

Time series are a type of time-dependent data found in many fields such as finance, medicine, meteorology, ecology, and the utility sector. In most of these fields, a time series may be dominated by several patterns, known as regimes \cite{sanquer2013smooth}, \cite{durichen2015multitask}. Identification of regimes allows for better handling of the behavioral variation of the series and improves time series forecasting. In a simple scenario, some series may periodically exhibit contiguous regimes over time. Fig. \ref{fig:contiguous regimes} \footnote{\scriptsize{\textit{Most of the figures used in the paper are best viewed in color/screen.}}} shows an example of a series dominated by three regimes, tagged respectively by blue, gray, and red boxes, which contiguously repeat over time. Such behavior can be seen for example in the utility sector, where a customer’s electricity consumption may be dominated by two main regimes, business week consumption and weekend consumption, which contiguously repeat over time. Series dominated by more than one regime that repeats over time are also known as multi-regime time series \cite{durichen2015multitask}.

To learn the behavioral variation of a time series such as the one illustrated in Fig. \ref{fig:contiguous regimes}, one can split the series into subseries each exhibiting a single regime, which, in fact, are easier to handle than multi-regime time series when attempting to forecast the series values at subsequent times \cite{tajeuna2018network}. Besides autoregressive \cite{tan2010day} approaches, neuronal \cite{cirstea2018correlated}, \cite{zhao2018forecasting} and other non-linear regression \cite{matsubara2016regime}, \cite{matsubara2016non},  \cite{wilson2013gaussian} methods are among the main learning models used for understanding series variation in the vast majority of applications. These methods are popular because they can capture non-stationarity due to the continuous changing behavior observed within the series. In practice, however, time series often include non-contiguous regimes, that is, regimes that do not follow a strict alignment. In such cases, it is difficult to determine how regimes are aligned compared to contiguous regimes as presented in Fig. \ref{fig:contiguous regimes}. 

\begin{figure*}[tbp]
\centering
\includegraphics[width=5.3in]{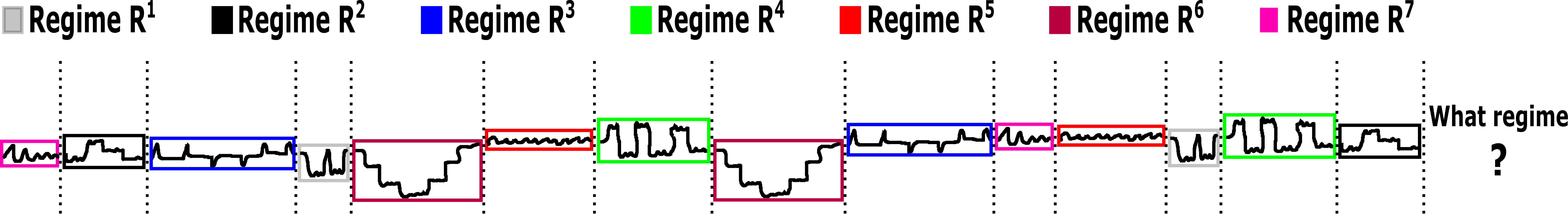}
\caption{ Series comprising seven non-contiguous regimes. Vertical dashed lines delimit regime time intervals.}
\label{fig:regime examples}
\end{figure*}

Fig. \ref{fig:regime examples} illustrates an example of a series containing seven regimes $R^1$ $-$ $R^7$, respectively identified by gray, black, blue, green, red, brown and pink colors in the figure. In this example, the regimes are non-contiguous. Such non-contiguous behavior is seen in many practical cases. For instance, in environmental studies, a gradual rise in atmospheric greenhouse gas concentrations can provoke abrupt climate transitions \cite{drijfhout2015catalogue}. In the area of health care, when patients with cognitive/heart problems are observed in their activities (e.g., walking, running, sitting), the signals collected may contain non-contiguous regimes whose discovery helps practitioners make further diagnostic and therapeutic decisions. In the financial area, unstable government policies and the mood of participants can influence the financial market and yield infrequent discount rate changes \cite{chen2012state}.

Although it could still be possible to split the series into subseries to allow a machine learning model to learn the series variation, the fact that regimes are non-contiguous causes a decision problem. For instance, at the $15^{th}$ time interval (the last one) in Fig. \ref{fig:regime examples}, it is difficult to know which regime will be applied, as one could for a series dominated by contiguous regimes. It is easy to see that directly training a nonlinear/neuronal model over such series without taking regime shifts into consideration may yield poor forecasting accuracy if a pre-analysis of the regime shift mechanism is not performed.

Recently, a number of methods \cite{dijk2002smooth}, \cite{matsubara2019dynamic}, \cite{cai2015fast}, \cite{deng2021pulse} addressing the forecasting of multiple time series that evolve together have tackled the issue of discovering latent regimes that series may exhibit. In the majority of these approaches, a model-based method is first used to handle the general regime shifts occurring within the multiple time series under investigation. The obtained model is then used to foresee the subsequent regimes that may be exhibited by the series at later time intervals. While advances have been made, the existing methods continue to suffer from shortcomings in one or more of the following areas:

\textbf{1) Regime identification:} In discovering regimes in multiple time series, existing methods usually do not consider the possible co-existence of series \cite{matsubara2019dynamic}, \cite{chen2018neucast}. In fact, the study of multiple time series as an ecosystem stipulates a co-evolutionary aspect between series which relates to the notion of complex systems and thus of possible interactions between these series at different time intervals. Exploring the interrelationships between series at different time intervals is crucial for regime identification and prediction.  

\textbf{2) Regime shifts:} Most of the existing methods attempt to capture the regime shift mechanism through a single transition matrix, which unfortunately cannot reflect the continuous transitions occurring in series dominated by non-contiguous regimes. An exception is the recent work of \cite{matsubara2019dynamic}, in which the authors define transitions between regimes over time. However, they did not study how long a regime may persist. The time persistence of a regime is indeed important in calculating the effective transition probability between regimes.

\textbf{3) Data discontinuity:} In real cases of co-evolutionary time series, some of the series may be discontinuous due to the absence of observations at particular time intervals. In the forecasting process, several existing methods either fill in the missing observations with estimated data or completely ignore them \cite{akccay2017short}, \cite{bokde2018novel}. This may bias the co-evolution of series and potentially lead to substantial loss of information. We hypothesize that discontinuity in a time series is an informative aspect of the series that needs to be taken into consideration rather than being ignored or replaced with virtual values that may potentially alter the series’ evolution.

\subsection{Overview of proposed model}
\begin{figure*}[tbp]
    \centering
    \includegraphics[width=5.5in]{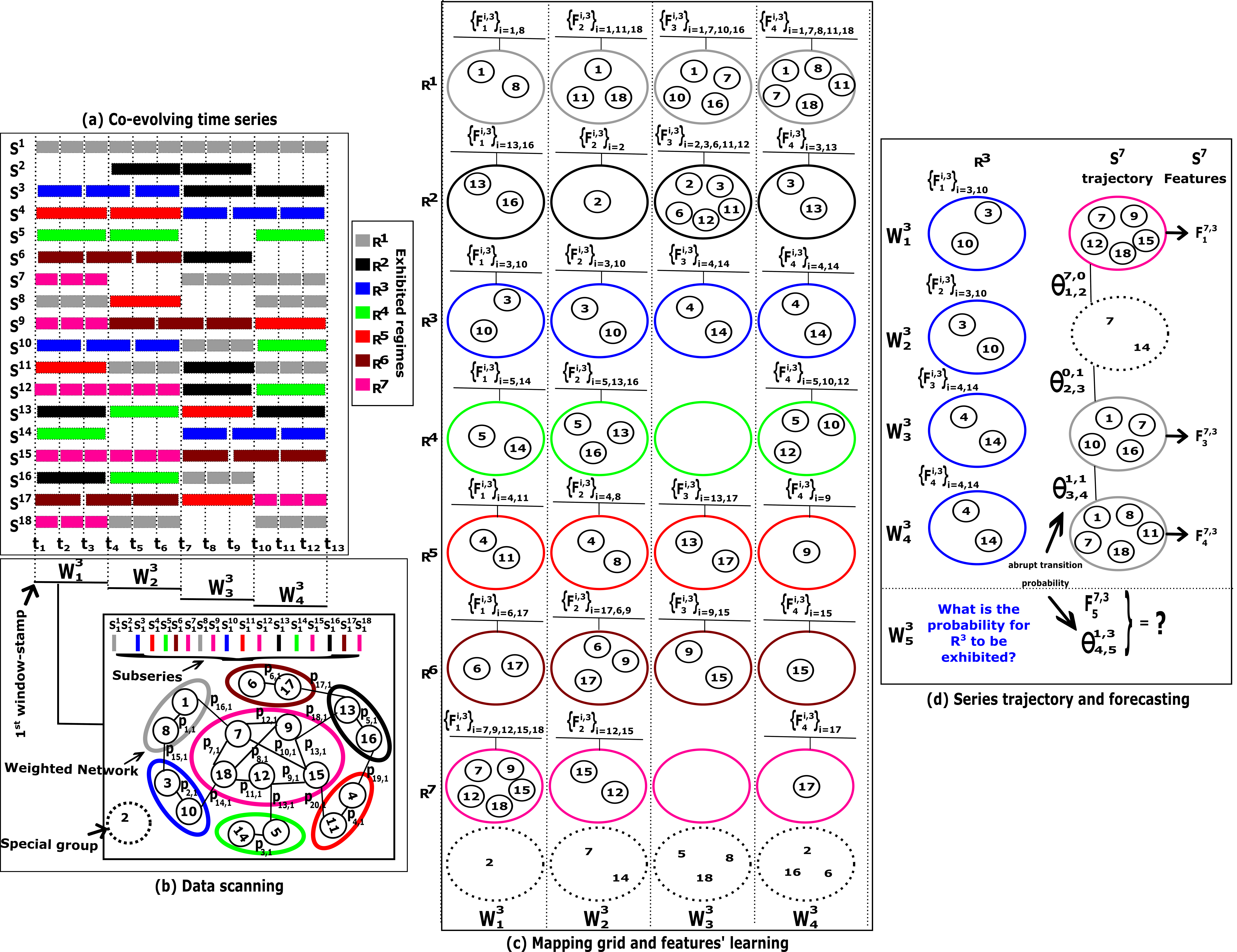}
    \caption{ Framework overview. $(a)$ shows $18$ time series exhibiting regimes $R^1$ $-$ $R^7$ within the time ranging from $t_1$ $-$ $t_{13}$. $(b)$ illustrates a data \textit{scanning} process with a window of size $3$ that slides from left to right. The \textit{scanning} process consists of building a weighted network ($\textbf{P}_{1,1}, \textbf{P}_{2,1}, ..., \textbf{P}_{20,1}$ are the weights of the edges at this first window-stamp $W^3_1$) structure from which patterns will be identified. $S^1_1$ $-$ $S^{18}_1$ are the subseries representing series $S^1$ $-$ $S^{18}$ respectively at the first window-stamp $W^3_1$. These subseries are represented by nodes ``$1$'' $-$ ``$18$'' of the weighted network. $(c)$ shows the mapping grid constructed using a sliding window $W^3$ and the learned features ($F^{i,\rho}_j$). Edges between nodes are ignored for purposes of visibility. Each row of the grid relates a regime evolution except the last row which records meaningful missing information. The columns summarize the distinct exhibited regimes at different window-stamps. $(d)$ depicts the trajectory followed by series $S^7$ and the transition probabilities from which it will be possible to forecast its values at window-stamp $W^3_5$. $F^{7,3}_5$ are the predicted features at the unknown  window-stamp $W^4_5$. $\Theta_{1,2}^{\alpha,\beta}$ $-$ $\Theta_{4,5}^{\alpha,\beta}$, $0<\alpha, \beta<7$ are the transition probabilities of a switch from one behavior to another.}
    \label{fig:scan process}
\end{figure*}

For the sake of clarity, consider the set of series depicted by Fig. \ref{fig:scan process}$(a)$. Here, we can see that regimes $R^1$ $-$ $R^7$ are exhibited by all series of the same data set at different time intervals. The observation of interchanging regimes across the whole data set shows that the series co-evolve over time. This is why possible interactions between series need to be considered in the process of regime identification. The observation of interchanging regimes in the data set further shows that most of the series continuously change behavior over time. Hence, defining a frozen matrix to account for the regime transition mechanism is not realistic since it cannot capture the temporal aspect of the regime shift. Furthermore, as can be seen, series are discontinuous due to missing data at different time points. As a regime is repeatedly exhibited, the repetitive gap observed at different time intervals can be informative. For instance, in series $S^2$, it could be said that after regime $R^2$ (tagged with black) is twice repeated, we have a three-unit gap of missing information. These recurrent gaps should be considered in the forecasting process as such, rather than replacing them with virtual values that may bias series evolution.

To address the aforementioned drawbacks of existing methods, we propose the following:

1) A watcher for regime identification. Under the assumption that co-evolving  time  series  are  dominated  by  a  finite  number  of  regimes having the same length, we start by automatically setting the size of a sliding window\footnote{More details about identifying the optimal window size  are given in Section \ref{article2-section:scanning data processus}.} from which we may obtain suitable regimes. Then, we slide the set window from left to right over the whole time series. Note that at each slide of the window (window-stamp), a \textit{scan} process is done. This process entails searching for homogeneous patterns via network clustering, as illustrated in Fig. \ref{fig:scan process}$(b)$. In this example, we have a \textit{window} of size $3$ that slides over time. At the $1^{st}$ slide, i.e. window-stamp $W^3_1$, the observed subseries $S^1_1$ $-$ $S^{18}_1$ respectively representing $S^1$ $-$ $S^{18}$ at $W^3_1$ are projected into a topological space, to form a network (or graph) structure. In this network, the nodes are labeled by indexes representing the observed subseries whereas the edges reflect the intensity of interaction or the similarity between the subseries through their weight values $p_{\zeta,1}$. The obtained network is then segmented into subnetworks each of which contains the subseries exhibiting a same regime.

2) Construction of a \textit{mapping grid} that depicts regime time duration. Such a \textit{mapping grid} enables an overview of regimes that have higher chance to be observed over time in the series. For illustration, in Fig. \ref{fig:scan process}$(c)$, we have for example a \textit{mapping grid} built from the window whose size was set to $3$ ($W^3$) in Fig. \ref{fig:scan process}$(b)$. In this example, each row $r$ ($1\leq r\leq 7$) of the grid illustrates the evolution of regime $R^r$. As can be seen, the number of nodes observed at each window-stamp of the seventh row (a number which gradually decreases), we may say that there is a high risk (probability) for regime $R^7$ to disappear. Rather than using such a naive approach to know whether a regime will no longer be observed or not, we apply a graph auto-encoder learning technique over the networks for learning features (sets $\{F^{i,3}_j\}$, $1\leq j \leq 4$ learned from each cell of a row in Fig. \ref{fig:scan process}$(b)$) that can better explain regime survival. 

3) Collection of historical data discontinuities via the sliding window. At each slide of the window, if there is a series for which no observation is found, that information is recorded in a special group. Note that the \textit{mapping grid} contains a specific row for collecting such information. In Fig. \ref{fig:scan process}$(c)$, this last row shows dashed circles containing indexes representing series for which no observation is found. For example, at the first slide of window $W^3$ (window-stamp $W^3_1$), it can be seen that there is no observation for series $S^2$. By collecting this information in the \textit{mapping grid}, it will be possible to foresee whether or not a series’ values will be observed at subsequent window-stamps.

By applying the strategy described above, the forecasting process becomes more effective. Given a time series belonging to a collection of co-evolving series, to forecast its values, we use the  \textit{mapping grid} to get its trajectory (i.e., its historical behavior), and then estimate the probability of its exhibiting a specific regime at subsequent window-stamps. In Fig. \ref{fig:scan process}$(d)$, for instance, we want to know the risk (probability) of regime $R^3$ being exhibited by series $S^7$ at window-stamp $W^3_{5}$. To this end, we first generate from the \textit{mapping grid} the trajectory followed by $S^7$. In the example, the trajectory is illustrated by the sequence of groups (vertical sequence of colored circles (on the right) in Fig. \ref{fig:scan process}$(d)$). From this sequence, we calculate the historical regime shift probabilities. In the example (Fig. \ref{fig:scan process}$(d)$), these probabilities are given by $\Theta_{1,2}^{\alpha,\beta}$ to $\Theta_{3,4}^{\alpha,\beta}$. After identifying the trajectory of series $S^7$, we then calculate the regime lifespan ($R^3$ in the example) using the sets of features $\{F^{i,3}_j\}$, $1\leq j \leq 4$, learned from the sequence of blue circles ($3^{rd}$ row of the mapping grid). When forecasting, we estimate the next feature values $F^{7,3}_5$ and the probability of shifting to regime $R^3$, $\Theta_{4,5}^{1,3}$, based on the historical features and the regime shift probabilities. The forecast regime that will be exhibited by series $S^{7}$ will correspond to the regime with the highest probability of being observed. Note that the transition probabilities $\bm{\Theta}^{\alpha,\,\beta}_{j,(j+1)}$ also capture the risk of having no observation at a window-stamp, as we can see in the $S^7$ trajectory. Here, from $W^3_1$ to $W^3_2$, the abrupt transition probability is given by $\bm{\Theta}^{7,\, 0}_{1,2}$.

\subsection{Contributions}
The significance of this work can be summarized as follows:

1) We propose a principled approach for multiple time series forecasting capable of identifying and handling hidden regime shifts exhibited by these series. The mapping grid devised in our approach is capable of capturing the overall continuous regime shift mechanism that occurs within the multiple time series data set.

2) We tackle the problem of continuous regime shifts by modeling a time-dependent probability of transition between regimes, using a full time-dependent Cox regression model. From the constructed mapping grid we calculate the survival of regimes and abrupt transition between regimes. The probability that a regime will be exhibited is calculated via a dynamic Cox regression model, whereas that of abrupt transition is calculated based on the constructed mapping grid.

3) We validate the whole approach by implementing it on both synthetic and real data sets. We illustrate the suitability of the proposed method by comparing its performance to that of six learning algorithms. Our experiments show that the proposed approach significantly improves forecasting accuracy.

\section{Related work}
In this section, we provide a high-level description of mainstream algorithms related to our work. While a complete survey is beyond the scope of this paper, we provide a critical review to put our work in perspective relative to existing methods. 

Smooth transition autoregressions are widely used for forecasting time series that exhibit various regimes. In some studies \cite{dijk2002smooth}, \cite{lubrano2000bayesian} it is first assumed that each exhibited regime can be recovered using an autoregressive model. Then, a weighted combination of these autoregressions is defined to foresee the series values at subsequent times. The linear combination is defined in a way that requires user intervention to manually fix the number of regimes. To sidestep this constraint, Sanquer et al. \cite{sanquer2013smooth} proposed a hierarchical Bayesian framework for automatically identifying the hidden regimes exhibited by series.

Note that in all of these methods, only one series can be handled at a time. To address the case of multiple time series, Hochstein et al. \cite{hochstein2014switching} proposed a multivariate smooth transition autoregression model in which each regime is modeled using a vector autoregressive model \cite{lutkepohl2005new}. In \cite{hochstein2014switching}, Hochstein also defined a higher-order Markov model to account for the regime shifts. This model is capable of capturing not only the most recent regime shift but also several previous regime shifts at once, using a single transition matrix. Though the plurality of series is considered in this case, the order of regimes and the order of the Markov model need to be manually set in advance.

In the methods mentioned above, it is important to note that the regime shifts are defined by means of a single matrix that states the overall transition probabilities between the exhibited regimes, meaning that the probability of switching from one regime to another remains constant. For series that exhibit non-contiguous regimes, however, the regime shift mechanism may be time-dependent. Moreover, the information gaps that certain series may present are not taken into consideration. To handle data discontinuity (due to the presence of missing information), Hallac et al. \cite{hallac2017network} proposed a network-based approach for tracking the evolution of co-evolving time series. Although their model can dynamically infer the graph structure, they did not address the problem of regime transitions. In \cite{matsubara2016regime}, the authors proposed the \textit{RegimeCast} model which learns, at a given window-stamp, the various patterns that may exist in a co-evolving environment, and reports the pattern(s) most likely to be observed at a subsequent time. Though the approach can report the subsequent patterns, it does not capture the possible dependencies that may exist between patterns. In updated work \cite{matsubara2019dynamic}, the authors proposed the deterministic \textit{ORBITMAP} model for capturing the time-dependent transitions between exhibited regimes. However, in their setting, they work with regimes that are priorly labeled (known in advance). Moreover, the \textit{ORBITMAP} model does not consider the case of data discontinuity. 

In \cite{li2009dynammo}, to overcome the data discontinuity problem, Li et al. proposed to complete missing values based on the neighboring trend followed by the data. Specifically, based on the observed part of the co-evolving time series, the authors in \cite{li2009dynammo} defined a sequence of latent variables that follow probabilistic rules. Then, based on the probabilistic model, they generated back series values at all time points including time intervals where we have no observation. Cai et al. \cite{cai2015fast} also proposed to solve the missing values problem by building the corresponding network structure relating to the interrelation between time series. The constructed network is later projected in a latent space through a matrix factorization from which the missing information gaps are no longer considered as a constraint since it is no longer perceived. In \cite{akccay2017short}, authors proposed an extended attention mechanism for reconstructing the existing gaps that may exist in time series whereas \cite{bokde2018novel} uses a forecast and backcast methodology to recover the missing values. However, in \cite{akccay2017short} and \cite{bokde2018novel}, the models are not made to handle multiple time series at once. Moreover, the methods proposed in \cite{li2009dynammo}, \cite{cai2015fast}, \cite{akccay2017short}, and \cite{bokde2018novel} were only for the cases where the time series have been accidentally harmed while being collected or manipulated, resulting in disappearance of values. However, in real cases, we may have multiple and repetitive missing values at different time intervals which are  instructive regarding series behavior and need to be taken as such.

Indeed, rather than ignoring or imputing the missing values, the authors in \cite{che2018recurrent}, \cite{habiba2020neural}, \cite{mikalsen2021time} investigate on how likely the missing gaps randomly appear and learn features from these gaps to fulfill their respective tasks. Despite recognizing the informativeness of missing values, the authors in \cite{che2018recurrent}, \cite{habiba2020neural}, \cite{mikalsen2021time} did not consider that gaps within the series may further appear and therefore influence in the series evolution. Instead, they rely on the current correlation of the gaps with respect to the series values to interpret how the series evolve. In our approach, we go beyond the current correlation between gaps and series values and further predict their subsequent appearances. 

Overall, in the current literature, various authors have contributed to advances in handling regime shifts in multiple time series, but only abrupt transition probabilities are captured by the models in all of these methods. In fact, some regimes may persist over time whereas others do not. Studying the time duration of regimes, or regime lifespan, is an important aspect that needs to be considered in the calculation of regime transition probability. To study event lifespan, statistical methods based on survival analysis have demonstrated their effectiveness in predicting failure/death events \cite{li2016multi}, \cite{li2017prospecting}, \cite{zhang2019time}. In this paper, we will use the Cox regression model for evaluating the risk of a regime being exhibited or not.

It is worth noting that, in the majority of applications of the Cox regression model, events are continuously tracked through a time-dependent baseline function which may be sensitive (continuously decreasing or increasing) to the effect of unchangeable covariates. However, due to the continuous changing regime shifts mechanism occurring within the time series ecosystem, using a Cox regression model with constant covariates will not well reflect the regime's survival probability. Facing this, we instead devise a full time-dependent Cox regression for modeling regime lifespans. In other words, we devise a Cox regression model where the covariates are time-dependent. 

\section{Key concepts}\label{article2-section:key concepts} 
In this section we describe key concepts and symbols used in our approach.

 \textbf{Multiple time series:} We call multiple time series an ensemble of univariate sequential values (a.k.a univariate series) that simultaneously evolve over regular time-stamps. We use $\bm{S} = \{S^i\}_{i=1}^N$ to denote a multiple time series comprising $N$ univariate time series. For the period ranging from time-stamp $t_1$ to $t_m$, $m>1$, each univariate time series $S^i \in \bm{S}$ is given as $S^i = \{e^i_l\}_{l=1}^m$ where $e^i_l$ is the value taken by series $S^i$ at time-stamp $t_l$. We use the set of indexes $\bm{\mathcal{N}} = \{i\}_{i=1}^N$ to respectively represent at all time points the univariate time series in $\bm{S}$.

\textbf{Sliding Window:} Given a set of time-stamps $\{t_l\}_{l=1}^m$, we call sliding window, a window of a given size that slides from left to right following the time-stamps $\{t_l\}_{l=1}^m$. We use the notation $W^{\rho}$ to denote a window of size $\rho$. To better illustrate, let us consider a window $W^3$ of size $3$ that slides from left to right. At the $1^{st}$ instance, this window covers the time-stamps $\{t_1,\,t_2,\,t_3\} = W^3_1$. At the $2^{nd}$ instance, the window covers the time-stamps $\{t_4,\,t_5,\,t_6\} = W^3_2$; etc. To generalize, for a window $W^{\rho}$ that slides, we use the notation $W^{\rho}_j$ to mean its $j^{th}$ instance, or $j^{th}$ window-stamp. The terms window-stamp and window instance will be used interchangeably throughout the paper to mean the same concept.

\textbf{Subseries:}
We use the term subseries to denote the sequence of values observed from a series at a given window instance. Formally, for a given series $S^i$, the corresponding subseries observed at the window instance $W^{\rho}_j$ is given as $S^{i,\; \rho}_j = \{e^i_l\}_{l = (j-1)*\rho+1}^{j*\rho}$. Hence, for a given sliding window $W^{\rho}$, we can split each series into contiguous subseries time segments of length $\rho$, as follows:

\begin{align}\label{article2-subseries}
    S^i &= \bigcup_{j=1}^{b} S^{i,\; \rho}_j
\end{align}

\noindent where $b = m/\rho$ is the number of instances of window $W^{\rho}$ in the time interval $[t_1,\,t_m]$.

\textbf{Time series network:}
From each set of subseries $\{S^{i,\;\rho}_j\}_{i=1}^N$ observed at window-stamp $W^{\rho}_j$, a \textit{time series network} is built as the projection of subseries $\{S^{i,\,\rho}_j\}_{i=1}^N$ onto a topological space $G_j = (\bm{\mathcal{N}},\, E_j,\, \mathcal{P}_j)$, a graph structure where $E_j$ and $\mathcal{P}_j$ are respectively sets of edges and edge weights that reflect the relationship strength between series (here represented by indexes $\bm{\mathcal{N}}$) at window-stamp $W^{\rho}_j$.

\textbf{Profile pattern:}
Given a set of subseries $\{S^{i,\;\rho}_j\}_{i=1}^N$ observed at window-stamp $W^{\rho}_j$. We can partition this set into homogeneous groups of subseries, $\{S^{i,\;\rho}_j\}_{i=1}^N = \{C^1_j, C^2_j, ..., C^r_j, ...\}$, where each group $C^r_j$ can be represented by a centroid $R^r$. We call profile pattern, the centroid exhibited by a set of homogeneous subseries. We use the notation $S^{i,\; \rho}_j\otimes R^r$ to mean that subseries $S^{i,\; \rho}_j$ exhibits the profile pattern $R^r$. Hence, at each window-stamp $W^{\rho}_j$, we then define a homogeneous set of subseries $C^r_j$ as $C^r_j = \{S^{i,\; \rho}_j\,|\,S^{i,\; \rho}_j \otimes R^r\}$.

It is important to note that, the different groups of homogeneous subseries at window-stamp $W^{\rho}_j$ can automatically be obtained by splitting the corresponding time series network $G_j$ into subnetworks $G^r_j = (\bm{\mathcal{N}}^r_j,\, E^r_j,\, \mathcal{P}^r_j)$ ($\bm{\mathcal{N}}^r_j \subseteq \bm{\mathcal{N}}$, $E^r_j \subseteq E_j$ and $\mathcal{P}^r_j \subseteq \mathcal{P}_j$) with nodes densely connected while being sparsely connected to other nodes of $G_j$. The set of nodes targeted by indexes $\bm{\mathcal{N}}^r_j$ corresponds again to subseries in $C^r_j$. In a nutshell, we can say that $G^r_j$ relates the group of subseries that exhibit the profile pattern $R^r$ projected in the topological space at $W^{\rho}_j$  whereas $C^r_j$ describes the group of subseries that exhibit the same profile pattern at $W^{\rho}_j$ in the time space.

\textbf{Regime:}
We call regime, a profile pattern observed at different window-stamps. Within the whole set of series $\bm{S}$, we assume that there exist a maximum of $K$ ($K>1$) possible regimes, denoted by $\bm{R} = \{R^r\}_{r=1}^K$. Note that, at a given window-stamp $W^{\rho}_j$, a regime $R^r$ might not be observed (i.e., there is no series exhibiting this regime). In this case, we have $C^r_j = \emptyset$. Analogous to the use of the label $R^r$, $1\leq r\leq K$ to denote existing regimes, we will use the label $R^0$ to denote no observation.

\textbf{Regime lifespan:} We use the term regime lifespan to denote the time period over which the regime continues to be observable. Noting that the regime is obtained from the group of subseries, the sole way of estimating the time duration of a regime is by tracking this group over time. The tracking is represented by a sequence of groups depicting the evolution of the regime. Formally, for a given regime $R^r$, the sequence representing its evolution over time (lifespan) is given as $\mathcal{E}^r = \{C^r_j\}_{j=1}^b$.

To investigate regime persistence over time, we utilize the graph representation (series projected in the topological space) for identifying and computing features. In fact, we use a Graph Autoencoder to automatically learn from the graph $G^r_j$ the embedding $\{F^{i,\rho}_j\}_{i\in \bm{\mathcal{N}}^r_j}$ that captures the series’ relationships while preserving their content \cite{wang2019attributed} (i.e., the subseries values). The obtained embedding is used as a knowledge feature that may help in understanding regime lifespan.

\begin{table}[tbp]
    \centering
    \caption{ Main notation.}
\label{table:notations}
    \begin{tabular}{l|l|l|l}
       Notation  & Definition & Notation & Definition \tabularnewline \hline
       $t_l$   & Time-stamp & $R^r$ & Regime \\
       $W^{\rho}$   & Window of size $\rho$ & $R^0$ & None regime \\
       $W^{\rho}_j$   & Window-stamp & $S^{i,\rho}_j\otimes R^r$ & $S^{i,\rho}_j$ exhibiting $R^r$ at $W^{\rho}_j$\\
       $S^i$   & Time series & $C^r_j$ & Group of subseries at $W^{\rho}_j$\\
       $\bm{\mathcal{N}}$  & Set of integers & $T^r$ & Profile pattern \\
       $S^{i,\rho}_j$   & Subseries at $W^{\rho}_j$ & $\mathcal{M}$ & Mapping grid \\
       $G_j$   & Network at $W^{\rho}_j$ & $\mathcal{E}^r$ & Regime lifespan \\
       $G^r_j$  & Subnetwork at $W^{\rho}_j$ & $Tr(S^i|W^{\rho}_b)$ & $S^i$ trajectory till $W^{\rho}_b$ \\
       $F^{i,\rho}_j$   & Node embedding at $W^{\rho}_j$ & $\Theta_{{j}\,,\,(j+1)}$  & Transition probability
    \end{tabular}
\end{table}

\textbf{Trajectory:} The term trajectory in our context refers to the changing behavior a given series displays over time: specifically, it corresponds to the different regimes the series exhibits. Here again, as in the definition of the regime lifespan, we recognize that a series may have missing values at certain window-stamps and use a sequence of groups to describe the various behavior displayed by a series over time. Formally, given a time series $S^i$ that evolves till window-stamp $W^{\rho}_b$, we define its trajectory as an ordered set or sequence

    \begin{align}\label{article2-trajectory}
        Tr(S^i\;|\;W^{\rho}_b) &= \bigcup_{j=1}^b\left\{R^{\alpha},\;0\leq \alpha \leq K \; |\; S^{i,\,\rho}_j\otimes R^{\alpha}\right\}
    \end{align}

\textbf{Transition probability:} The transition probability is defined as the risk of series $S^i$ switching from a regime $R^{\alpha}$, $\alpha \in \{0,\, 1,\, 2,\,...\,K\}$ it exhibits at window-stamp $W^{\rho}_j$ to another regime $R^{\beta}$, $\beta \in \{0,\, 1,\, 2,\,...\,K\}$ at window-stamp $W^{\rho}_{j+1}$. In this paper, we use the symbol $\bm{\Theta}_{{j}\,,\,{(j+1)}}$ to refer to the matrix of transition probabilities from one regime to another between two consecutive window-stamps $W^{\rho}_j$ and $W^{\rho}_{j+1}$.

To conclude this section, Table \ref{table:notations} provides the main notation used throughout the paper.

\section{Proposed approach}\label{article2-proposed approach}
To model and forecast regimes shifts in multiple time series, we propose an approach that proceeds in three phases: (1) scanning the input multiple time series for identifying repetitive profile patterns, (2) restructuring the multiple time series into a grid that maps the series behaviors and regime lifespans and (3) building a survival model for multiple time series forecasting. Details of each phase are given in the following.

\subsection{Phase 1: Data scanning}\label{article2-section:scanning data processus}
Given an ensemble of multiple time series, the main idea behind the data scanning step is to apply a network-based approach for discovering regimes that series may exhibit over time. To do this, we need to set the size of a suitable window from which we observe how the series evolve conjointly and identify the regimes that will be represented by the profile of similar series at each time stamp.

To set the window size, we use an incremental procedure in which we test various window sizes through a process of data scanning until an optimal one is obtained. Given a set of window sizes $\mathcal{D}$, for each value $\rho \in \mathcal{D}$, we define a window $W^{\rho}$ which slides from left to right, scanning the series in the aim of identifying significant repetitive patterns. At the end of the sliding process, i.e., after passing through the series with the window, we define a \textit{regime-density-per-window-size} score which will help us determine whether we have reached the optimal window size.

\begin{figure}[tbp]
    \centering
    \includegraphics[width=5in]{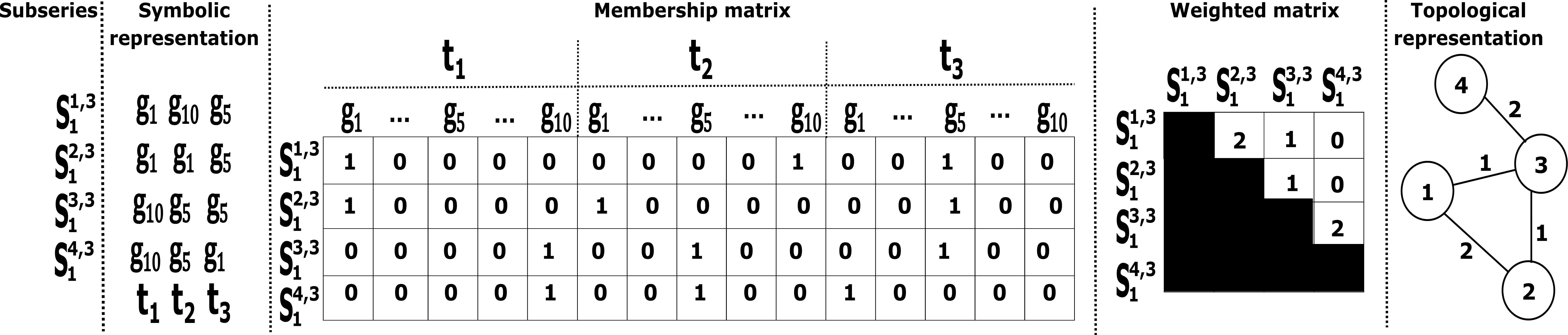}\vspace{-2mm}
    \caption{ Process flow for network construction. $S^{i,3}_1$ are subseries extracted at window-stamp $W^3_1$. The weight matrix is symmetric; thus, only the upper or lower triangle need be considered. Integers $1$ $-$ $4$ represent subseries $S^{1,3}_1$ $-$ $S^{4,3}_1$, respectively, in the topological representation. Edge weights are values from the weight matrix.}
    \label{fig:weighted graph}\vspace{-3mm}
\end{figure}

\subsubsection{Network construction}\label{article2-section:network construction}

Given a window $W^{\rho}$, based on Eq.(\ref{article2-subseries}) at each window-stamp $W^{\rho}_j$, we build a network from the set of subseries derived from the series found in $\bm{S}$. 

The entities represented by nodes are subseries, while interrelations are represented by edges. 

To capture the relationship between subseries, we make use of the Symbolic Aggregate approXimation (SAX) \cite{lin2007experiencing} to discretize each of our subseries into categorical data \cite{senin2018grammarviz}. Specifically, knowing that the time series are scaled within the interval $[0,\,1]$, we uniformly codify values of this interval using a $10-symbol scale = \{g_a\}_{a=1}^{10}$, where the symbol $g_a$ is considered as a random variable which can only take values in $[(a-1)/10,\;a/10)$. Thus, at each window-stamp $W^{\rho}_j$, each subseries can now be rewritten as a sequence of symbols. Having the symbolic representation of our subseries, we now build a membership matrix $\bm{\Delta}_j = (\delta^i_{j,\tau})$, $1\leq i\leq \bm{\mathcal{N}}$, $1\leq \tau \leq 10\,\rho$ where columns represent symbols and rows correspond to subseries. We use the values $1$ and $0$, respectively, to denote the presence and the absence of the symbol in the discretized series. Based on the membership matrix $\bm{\Delta}_j$, we then calculate the weighted matrix $\mathcal{P}_j = Upper(\bm{\Delta}_j \times \bm{\Delta}^*_j$, where $\bm{\Delta}^*_j$ corresponds to the transpose of matrix $\bm{\Delta}_j$ and $Upper(\bm{\Delta}_j \times \bm{\Delta}^*_j)$ the upper triangle of matrix $\bm{\Delta}_j \times \bm{\Delta}^*_j$. Fig. \ref{fig:weighted graph} provides an illustration of the procedure we implemented to build the network reflecting the relationship between subseries $S^{1,3}_1$ $-$ $S^{4,3}_1$.

\subsubsection{Window size selection}\label{article2-section:series variation and window selection}
At each window-stamp $W^{\rho}_j$, we construct the network $G_j$, and then use a network-based clustering algorithm to split $G_j$ into subnetworks of closely connected nodes \cite{boutemine2017mining}. In our experiments, we used Infomap \cite{rosvall2008maps} because it is an effective parameter-free algorithm capable of clustering a networked data without any parameter setting as Lancichinetti et al. demonstrate in \cite{lancichinetti2009community}. It is important to note that, at each window-stamp $W^{\rho}_j$, the number of identified subnetworks may vary. Our aim is to find an optimal window size permitting highly similar, or repetitive, subnetworks across the networks at different window-stamps. The profile patterns of such subnetworks are defined as the regimes. The repetitiveness allows generating similar regimes at different window-stamps for tracking and lifespan analysis. We propose a heuristic solution to this problem via the segmentation of a score function that is defined as follows:

    \begin{align}\label{article2-series_dynamic}
        \mathcal{V}ar(\rho\,|\,\bm{S}) &=  \frac{1}{\rho}\cdot max \left\{K_j,\; 1 \leq j \leq b\right\}   
    \end{align}

\noindent where $K_j$ is the number of subnetworks identified at window-stamp $W^{\rho}_j$, which means that, $max \left\{K_j,\; 1 \leq j \leq b\right\}$ corresponds to the maximum number of regimes observed in $\bm{S}$.

\begin{figure}[tbp]
    \centering
    \includegraphics[width=2.35in]{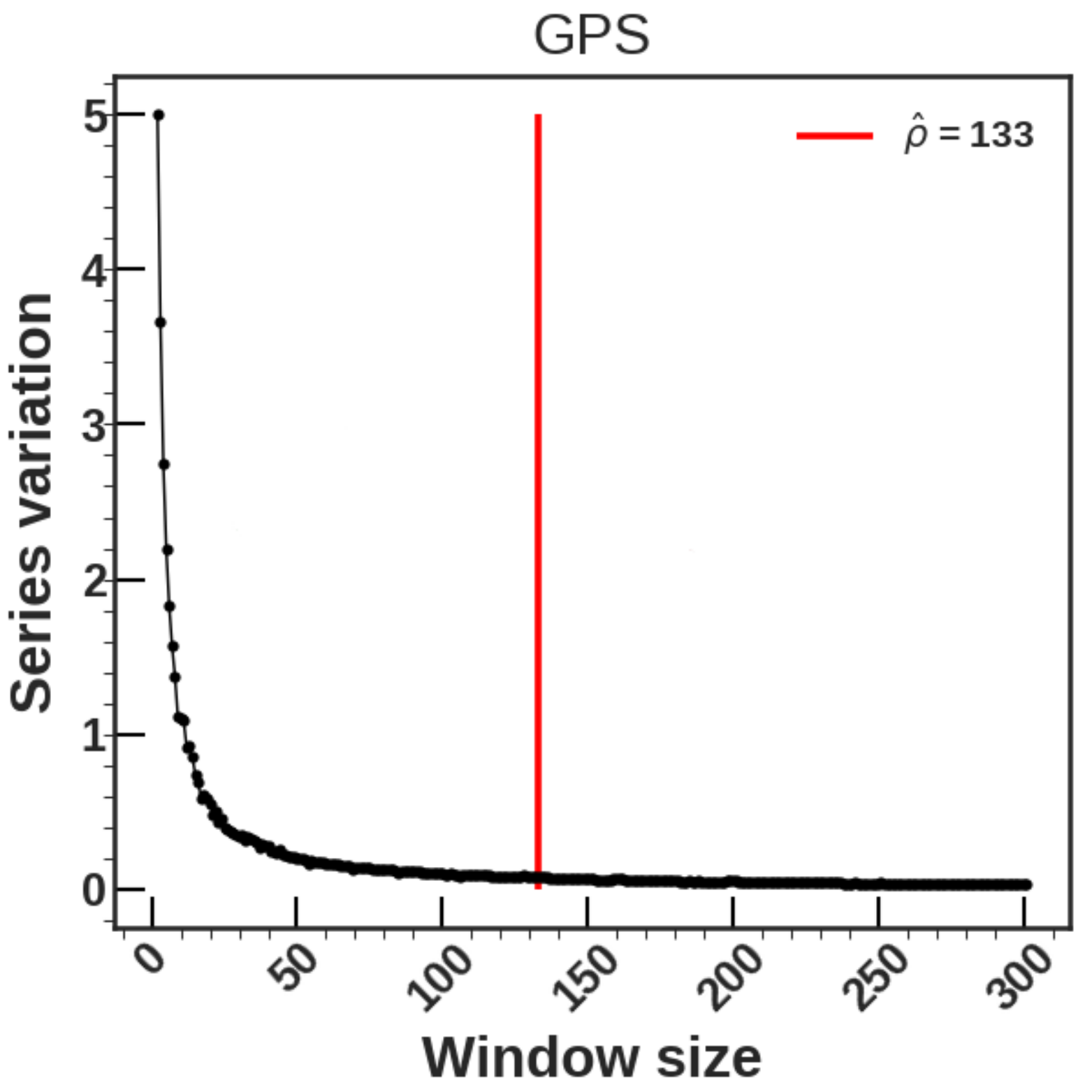}
    \caption{
    Example of trends followed by $\mathcal{V}ar()$ in the gesture phase segmentation data set (GPS). The vertical line corresponds to the border between the large and small values. The window size corresponding to this border is selected as the optimal window size $\hat{\rho}$.}
    \label{fig:convex function}
    \vspace{-3mm}
\end{figure}

Eq. (\ref{article2-series_dynamic}) defines a \textit{regime-density-per-window-size} score that explains how the whole time series varies with respect to the size of the window. Intuitively and experimentally, we have the following observations. A narrow window size $\rho$ leads to high regime density score $\mathcal{V}ar()$, which corresponds to the cases where there exist regimes that are not repeating over time: the data set is unstable when looking through a narrow window. On the other hands, a wide window size $\rho$ leads to low regime density score $\mathcal{V}ar()$, which corresponds to the cases with highly repetitive regimes. Hence, it is important to select a window which is wide enough for finding suitable regimes. However, $\mathcal{V}ar()$ is a decreasing function that converges to the null density. A null density means that the window size is too large to permit capturing regime shift events. To avoid being in such situation, we select the smallest window size from which we can observe the quasi-stability in terms of regimes that repeat over time. 

To automatically select the optimal window size $\hat{\rho}$, we use the MDL pruning techniques \cite{agrawal1998automatic}, \cite{bouguessa2008mining} over the set of regime density scores $\bm{\mathcal{V}ar} = \{\mathcal{V}ar(\rho\,|\,\bm{S}),  \rho \in \mathcal{D}\}$. In \cite{agrawal1998automatic}, \cite{bouguessa2008mining}, the authors used MDL selection to split their data into two subsets (sparse and dense subsets) and then discard one of the two. In our case, we are selecting the border between the two subsets from which we can get the window size that minimizes the MDL criteria, as follows:

    \begin{align}\label{article2-eq:optimal_window_size}
        \hat{\rho} &= \begin{cases}
                \underset{\rho \in \mathcal{D}}{argmin}\{CL(\rho)\}\\
                CL(\rho) = log_2(\overline{\bm{\mathcal{V}ar_1}}) + \\ 
                \;\;\; \sum_{\rho \in \mathcal{D}_1}log_2(\lvert (\mathcal{V}ar(\rho|\bm{S}) - log_2(\overline{\bm{\mathcal{V}ar_1}})\rvert) + \\
                \;\;\; log_2(\overline{\bm{\mathcal{V}ar_2}})+ \sum_{\rho \in \mathcal{D}_2}log_2(\lvert (\mathcal{V}ar(\rho|\bm{S})  - log_2(\overline{\bm{\mathcal{V}ar_2}})\rvert)
            \end{cases}
    \end{align}
\normalsize
\noindent where $\mathcal{D}_1=\mathcal{D}\setminus \{\rho+1,\,...,\,m-1\}$, $\mathcal{D}_2 = \mathcal{D}\setminus \{2,\,...,\,\rho\}$, $\bm{\mathcal{V}ar_1} = \{\mathcal{V}ar(\rho\,|\,\bm{S}),\; \rho \in \mathcal{D}_1\}$, $\bm{\mathcal{V}ar_2} = \{\mathcal{V}ar(\rho\,|\,\bm{S}),\; \rho \in \mathcal{D}_2\}$ and $\overline{\bm{\mathcal{V}ar_1}}$ (resp. $\overline{\bm{\mathcal{V}ar_2}}$) the average variation score in $\mathcal{D}_1$ (resp. $\mathcal{D}_2$). The selected window size is taken as the optimal window size  $\hat{\rho}$. See Fig. \ref{fig:convex function} for an illustration.

Having estimated the optimal window size $\hat{\rho}$, we then use this size to get the number of distinct regimes $K$ by counting the number of distinct profile patterns across all window-stamps $W^{\hat{\rho}}_j$, $1\leq j \leq b$.


\begin{algorithm}[tbp]
\small{
\caption{Data scanning \& grid construction}
 \label{algo:down drill}
    \KwData{$\bm{S} = \{S^i\}_{i=1}^N,\; \mathcal{D},\; \epsilon$ \scriptsize{\tcc*{Sets of co-evolving time series, window sizes and convergence criteria.}}}
    \KwResult{$\hat{\rho},\;K,\; \bm{SS},\;\bm{\mathcal{V}ar},\; \mathcal{M}$ \scriptsize{\tcc*{Optimal window size, number of regimes, set of regimes, the series dynamic and the mapping grid.}}}  
    \Begin{
        \textbf{A- Data scanning:} \\
        \Begin{
            \textbf{A.1- Initialization:}\\
            \Begin{
                - $\rho \gets $ First value in $\mathcal{D}$\;
                - $\bm{\mathcal{V}ar} \gets \emptyset$\scriptsize{\tcc*{Initialize the set of scores.}}
            }
            \textbf{A.2- Scanning:} \\
            \Begin{
                - At each window-stamp $W^{\rho}_j$ use Eq.(\ref{article2-subseries}) to get the set of subseries $\{S^{i,\rho}_j\}_{i=1}^N$\;
                - Use $\{S^{i,\rho}_j\}_{i=1}^N$ to Build the network $G_j$ as discussed in section \ref{article2-section:network construction}\;
                - Split $G_j$ in to subnetworks and get the set of groups $\mathcal{O}_j$ at each window-stamp $W^{\rho}_j$ as discussed in section \ref{article2-section:series variation and window selection}\;
                - Using Eq. (\ref{article2-series_dynamic}), calculate the \textit{regime-density-per-window-size} $\mathcal{V}ar(\rho\,|\,\bm{S})$\;
                - $\bm{\mathcal{V}ar} \gets \bm{\mathcal{V}ar} \bigcup \{\mathcal{V}ar(\rho\,|\,\bm{S})\}$\;
            }
            \textbf{A.3- Iteration:} \\
            \Begin{
                \While{True}{
                    - $\rho \gets $ next value in $\mathcal{D}$\;
                    - Using the new value $\rho$ scan series by running step \textbf{A.2} and get the $new \; \mathcal{V}ar(\rho\,|\,\bm{S})$\;
                    - If $\lvert new \; \mathcal{V}ar(\rho\,|\,\bm{S}) - previous \; \mathcal{V}ar(\rho\,|\,\bm{S})\rvert < \epsilon$ stop the loop\;
                }
            }
            - Based on the set of scores $\bm{\mathcal{V}ar}$ get the optimal window $\hat{\rho}$ as discussed in section \ref{article2-section:series variation and window selection}\;
            - $\bm{SS} \gets $ distinct regimes, as mentioned in section \ref{article2-section:series variation and window selection} and $K \gets \lvert \bm{SS}\rvert$\;
        }
        \textbf{B- Mapping grid:} \\
        \Begin{
             - Initialize a $(K+1)\times b$ empty mapping grid $\mathcal{M}$, where $b$ is the number of instances of $W^{\hat{\rho}}$\;
             - At each window-stamp $W_j^{\hat{\rho}}$, populate cells of the mapping grid with indexes $\mathcal{N}^r_j$ and $\mathcal{N}_j$ representing series exhibiting each regime $R^r$ and series from which we have no observation respectively\;
             - For each cell of the mapping grid, learn the corresponding features as discussed in section \ref{article2-section:grid and features extraction} 
        }
    }
}
\end{algorithm}

\subsection{Phase 2: Mapping grid construction \& regime survival}\label{article2-section:mapping grid construction processus}
In this section, we first present how to build and populate the mapping grid. Then, based on the constructed grid, we present how to learn features, i.e., descriptors that may explain regime persistence, that will be used to predict regime survival. Finally, we illustrate how to estimate the regime survival probability using the learned features. 

\subsubsection{Grid construction and features learning}\label{article2-section:grid and features extraction}

As shown in Fig. \ref{fig:scan process}$(c)$, the mapping grid is designed as a $(K+1) \times b$ grid $\mathcal{M}$ of nodes each representing a group of subseries exhibiting a regime. The mapping grid provides  an  overview  of regimes persisting  or  disappearing  over  time and allows extraction of the trajectory of each series. In fact, with the size of the window set to $\hat{\rho}$ and the number of regimes to $K$, we determine which regime a series exhibits at window-stamp $W^{\hat{\rho}}_j$, $1\leq j \leq b$. This also includes the case of no regime when there is no observation. The last row of $\mathcal{M}$ contains indexes indicating those series for which there was no observation. For each regime $R^r$, we can get from the grid the sequence $\mathcal{E}^r = \{C^r_j\}_{j=1}^b$ of groups of subseries representing the regime lifespan. 

From the constructed mapping grid, features need to be learned in order to better explain the regime lifespan by considering the interrelation between subseries of the same group. To take these interdependencies into consideration, we propose to use a graph autoencoder to automatically learn features from each node of our mapping grid. The principle behind the graph autoencoder consists in representing a network structure into a small latent space while preserving node interrelations and being capable of approximately reconstructing the network using the latent representation \cite{tu2017transnet}, \cite{hamilton2017representation}. Dimensions of the latent representation (the embedding) are used as the network features. The different steps for building the mapping grid from a set of co-evolving time series is summarized in Algorithm \ref{algo:down drill}. 

\subsubsection{Regime survival}\label{article2-section:survival probability}

Now that we have constructed a grid capable of capturing various regime behaviors exhibited by series over time, we turn to learning the probabilities with which these series change behaviors over time. Before addressing the regime shift occurring in each series, we first need to learn each regime's lifespan. Given a regime $R^r$, at each instance $j$ of the window $W^{\hat{\rho}}$, we count the number of series $\mathcal{J}^r(j)$ exhibiting this regime, as follows:

	\begin{align}\label{article2-counting process1}
		\mathcal{J}^r(j) &= H^r(j) + M^r(j)
	\end{align}

\noindent where $H^r()$ is a non-decreasing function, called \textit{cumulative intensity process}, and $M^r()$ is a mean-zero martingale \cite{aalen2008survival}. 

Considering $H^r()$ to be continuous, there exists a predictable non-negative intensity process $h^r()$ such that

       \begin{align}\label{article2-counting process2}
       	H^r(j) &= \int_{\omega = 1}^{\omega = j} h^r(\omega)d\omega
       \end{align}

At each window-stamp $W^{\hat{\rho}}_j$, we calculate the hazard for a regime $R^r$ to be observed as:

\begin{align}\label{article2-Risk on event}
    h^r(j) &= \begin{cases}
        0\;\;\;\; if \;\;\;\;\lvert C^r_j \rvert = 0\\
        \frac{\gamma^r(j)}{\lvert \bm{\mathcal{N}}^r_j \rvert} \sum_{S_j^{\,i,\,\hat{\rho}} \in C^r_j} exp(F_j^{\,i,\,\hat{\rho}}\; \bullet\; \Gamma_j^r)
        \;\;\;\; otherwise
    \end{cases}
\end{align}

\noindent  where the ``$\bullet$'' symbol stands for the dot product, $\Gamma_j^r$ the covariates representing the contribution of the feature $F_j^{\,i,\,\hat{\rho}}$ at $W^{\hat{\rho}}_j$. This vector parameters are learnt based on the historical information collected till window-stamp $W^{\hat{\rho}}_j$. $\gamma^r()$ is a Gamma probability density function that plays the role of the baseline hazard. Note that, any probability density function that fits with the distribution of the number of subseries exhibiting the given regime can be considered. Our choice has gone to the Gamma function due to its great shape flexibility. 
 
Knowing the historical features till the actual feature $\{F_j^{\,i,\,\hat{\rho}}\}_{i\in\bm{\mathcal{N}}^r_j}$, we have a partial view of the data ensemble. For this reason, we define the partial log-likelihood of parameters $\Gamma_j^r$ over the observed feature values \\ $\left\{\{F_l^{\,i,\,\hat{\rho}}\}_{i\in\bm{\mathcal{N}}^r_l},\; 1\leq l\leq j\right\}$ as follows: 

    \begin{align}\label{article2-partial_loglikelihood}
        \mathcal{LL}(\Gamma_j^r) &= \sum_{l=1}^{j-1} \sum_{S_l^{i,\hat{\rho}} \in C^r_l} \nu^r_l \cdot \mathcal{L}\left(\frac{ exp(F_l^{i,\hat{\rho}} \bullet \Gamma_l^r)}{\sum_{S_l^{i,\hat{\rho}} \in C^r_l} exp(F_l^{i,\hat{\rho}} \bullet \Gamma_l^r)}\right)
        \notag \\
        &= \sum_{l=1}^{j-1} \sum_{S_l^{i,\hat{\rho}} \in C^r_l} \nu^r_l \left[(F_l^{i,\hat{\rho}} \bullet \Gamma_l^r ) - \mathcal{L}\left(\sum_{S_l^{i,\hat{\rho}} \in C^r_l} exp(\vec{F}_l^{i,\hat{\rho}} \bullet \vec{\Gamma}_l^r) \right)\right]
    \end{align}

\noindent where $\nu^r_l$ is a binary indicator which takes $0$ if $C^r_l = \emptyset$ and $1$ otherwise and $\mathcal{L}$ the logarithme.

From the above partial log-likelihood, we learn the parameter $\Gamma_j^r$ by using the Newton-Raphson algorithm. The process is repeated each time the window slides. In this way, we always have all the parameters $\Gamma_j^r$ updated with time. From Eq.(\ref{article2-Risk on event}) we can then calculate, the lifespan probability of a regime till the $j^{th}$ instance of window $W^{\hat{\rho}}$, by computing the cumulative probability $Surv(j\,|\,R^r)$ \cite{aalen2008survival}, as follows 

       \begin{align}\label{article2-Survival relation}
       Surv(j\,|\,R^r) &= exp\Bigl\{-\int_{\omega = 1}^{\omega = j}h^r(\omega)d\omega \Bigr\} 
       \end{align}

\subsection{Phase 3: Multiple time series forecasting}\label{article2-section:series forecasting}

Given $\bm{S}$, a set of co-evolving time series, now that we have constructed our mapping grid, in order to foresee the series values at a subsequent time, we need to know about regime transition, from which we will be able to forecast the subsequent values of series in $\bm{S}$. 

\subsubsection{Regime transition}\label{article2-matrix transition}

From the mapping grid $\bm{\mathcal{M}}$, for two consecutive window-stamps $W^{\hat{\rho}}_j$ and $W^{\hat{\rho}}_{j+1}$ we can calculate the probabilities $\mathcal{Q}_{j,(j+1)}$ of switching from behavior $R^{\alpha}$ to behavior $R^{\beta}$ by using a counting process as follows: 

    \begin{align}\label{article2-sudden transition}
        \mathcal{Q}_{j,(j+1)} &=  \left(\frac{\lvert \bm{\mathcal{N}}^{\alpha}_j \bigcap \bm{\mathcal{N}}^{\beta}_{j+1} \rvert}{\lvert \bm{\mathcal{N}}^{\alpha}_j \bigcup \bm{\mathcal{N}}^{\beta}_{j+1} \rvert}\right)_{\alpha,\; \beta \in \{0,\,1,\,...,\,K\}} \notag \\
        &= \left(\mathcal{Q}^{\alpha,\,\beta}_{j,\,(j+1)}\right)_{\alpha,\; \beta \in \{0,\,1,\,...,\,K\}}
    \end{align}

\noindent where $\bm{\mathcal{N}}^{\alpha}_j$ (resp. $\bm{\mathcal{N}}^{\beta}_j$) is the number of series presenting an $R^{\alpha}$ (resp. $R^{\beta}$) behavior at window-stamp $W^{\hat{\rho}}_j$.

Note that the component $\mathcal{Q}^{\alpha,\,\beta}_{j,\,(j+1)}$ again identifies the instantaneous risk/ probability of observing an $R^{\beta}$ behavior while knowing that we previously had an $R^{\alpha}$ behavior. In other words, the matrix $\mathcal{Q}_{j,(j+1)}$ depicts the probability of suddenly switching from one behavior to another within the overall data set $\bm{S}$. However, if we want to calculate the probability of suddenly switching from one behavior to another for a single series $S^i \in \bm{S}$ we cannot uniquely exploit the overall risk phenomenon calculated in Eq. (\ref{article2-sudden transition}). 

In the absence of other series, for a specific series $S^i$, we believe that the risk of suddenly observing an $R^{\beta}$ behavior knowing that the previous exhibited behavior was $R^{\alpha}$ is calculated according to the series’ historical behaviors, as follows: 

    \begin{align}\label{article2-sudden transition2}
        \Pi(W^{\hat{\rho}}_j,W^{\hat{\rho}}_{j+1}\,|\,S^i) &=  \left(\frac{\eta\left(R^{\alpha}\rightarrow R^{\beta}|\mathcal{T}r(S^i\,|\,W^{\hat{\rho}}_{j+1})\right)}{|\mathcal{T}r(S^i\,|\,W^{\hat{\rho}}_{j+1})|-1}\right)_{\alpha, \beta \in \{0,\,1,\,...,\,K\}}\notag \\
        &= \left(\Pi^{\alpha,\,\beta}_{j,\,(j+1)}\right)_{\alpha,\; \beta \in \{0,\,1,\,...,\,K\}}
    \end{align}

\noindent where $\eta\left(R^{\alpha}\;\rightarrow\; R^{\beta}\;|\;\mathcal{T}r(S^i\;|\;W^{\hat{\rho}}_{j+1})\right)$ is the number of time the sequence $\{R^{\alpha},\;R^{\beta}\}$ appears in the trajectory $\mathcal{T}r(S^i\;|\;W^{\hat{\rho}}_{j+1})$ and $|\mathcal{T}r(S^i\;|\;W^{\hat{\rho}}_{j+1})|$ the length of the trajectory.

With the relations given in Eq. (\ref{article2-sudden transition}) and Eq. (\ref{article2-sudden transition2}), we can take into consideration the existence of other series in $\bm{S}$ and thus calculate the effective probability $\bm{\Theta}(W^{\hat{\rho}}_j,W^{\hat{\rho}}_{j+1}\,|\,S^i) = \left(\bm{\Theta}^{\alpha,\,\beta}_{j\; ,\; (j+1)}\right)_{\alpha,\; \beta \in \{0,\,1,\,...,\,K\}}$ of series $S^i$ suddenly switching from behaviors $R^{\alpha}$ to behaviors $R^{\beta}$, as follows: 

    \begin{align}\label{article2-sudden transition3}
        \bm{\Theta}^{\alpha,\,\beta}_{j\; ,\; (j+1)} &= \begin{cases}
            0 \;\;\;\; if \;\;\;\; \sum_{\star_1}\sum_{\star_2}\Pi^{\star_1,\,\star_2}_{j,\,(j+1)}\mathcal{Q}^{\star_1,\,\star_2}_{j,\,(j+1)} = 0\\
            \frac{\sum_{\star}\Pi^{\alpha,\,\star}_{j,\,(j+1)}\mathcal{Q}^{\star,\,\beta}_{j,\,(j+1)}}{\sum_{\star_1}\sum_{\star_2}\Pi^{\star_1,\,\star_2}_{j,\,(j+1)}\mathcal{Q}^{\star_1,\,\star_2}_{j,\,(j+1)}} \;\;\;\; if \;\;\;\; not
        \end{cases}
    \end{align}

It is worth noting that each component $\bm{\Theta}^{\alpha,\beta}_{j\; ,\; (j+1)}$ is a conditional probability that gives the risk of suddenly observing an $R^{\beta}$ behavior in a given series $S^i \in \bm{S}$, knowing that this series was previously exhibiting an $R^{\beta}$ behavior.

\subsubsection{Forecasting}\label{article2-forecasting series} 

For each time series $S^i$ observed from window-stamp $W^{\hat{\rho}}_1$ to $W^{\hat{\rho}}_b$, we want to predict the regime this series will exhibit at the next window-stamp $W_{b+1}^{\hat{\rho}}$. To this end, we start by predicting the general switching $\mathcal{Q}_{b,(b+1)}$, the conditional transition $\Pi(W^{\rho}_b,\, W^{\rho}_{b+1}\;|\;S^i)$ and the feature vector values $\{\vec{F}^{i,\hat{\rho}}_{b+1}\}_{i\in\bm{\mathcal{N}}^r_{b+1}}$, $1\leq r\leq K$. The predicted matrices $\mathcal{Q}_{b,(b+1)}$ and $\Pi(W^{\rho}_b,\, W^{\rho}_{b+1}\;|\;S^i)$ are used for calculating the effective probability of suddenly changing $\bm{\Theta}(W^{\hat{\rho}}_b,W^{\hat{\rho}}_{b+1}\,|\,S^i)$ at window-stamp $W^{\hat{\rho}}_{b+1}$ by applying Eq. (\ref{article2-sudden transition3}). The predicted feature vector values are used for calculating the regime survival till window-stamp $W^{\hat{\rho}}_{b+1}$ by applying Eq. (\ref{article2-Survival relation}). It is worth noting that, to predict the transition matrices values as well as the feature vector values, we use the regression form given as follows:

    \begin{align}\label{article2-generate features}
        A_{b+1} &= \lambda(A_1,\,...,\,A_b) + \mu_{b+1}
    \end{align}

\noindent where $A_b$ could be a matrix or a feature vector at window-stamp $W^{\hat{\rho}}_b$.\\ $\lambda(A_1,\,...,\,A_b)$ a given regression function and $\mu_{b+1}$ a white noise following a zero mean Gaussian function. 

Knowing the survival $Surv(W^{\rho}_{b+1}|R^r)$ of regime $R^r$ and the conditional transition $\bm{\Theta}^{\alpha,\beta}_{b\,,\,(b+1)}$ of series $S^i$, we can then forecast the series $S^i$ values at window-stamp $W^{\hat{\rho}}_{b+1}$. If the most probable transition goes to one of regimes $R^r$, $1\leq r \leq K$, then the predicted values $\widehat{S^{\;i,\; \hat{\rho}}_{b+1}}$ are calculated as:

\begin{align}\label{article2-forecasted values}
\widehat{S^{\;i,\;\hat{\rho}}_{b+1}} &= R^{r} + \left[\overline{R}^{\,r} \pm \frac{\left(1 - \widehat{\mathcal{TP}}_{R^{\alpha} \rightarrow R^r}(W^{\hat{\rho}}_{b+1})\right)\cdot \sigma(R^{r})}{2}\right]
\end{align}

\noindent where $\overline{R}^{\,r}$ is the mean value of the profile pattern $R^{r}$, $\sigma(R^{r})$ its standard deviation. $\widehat{\mathcal{TP}}_{R^{\alpha} \rightarrow R^r}(W^{\hat{\rho}}_{b+1})$ is the highest transition probability for $S^i$ (corresponding to the regime it is most likely to switch to) given as:
\begin{align}\label{article2-highest transition}
    \widehat{\mathcal{TP}}_{R^{\alpha} \rightarrow R^{\beta}}(W^{\hat{\rho}}_{b+1}) &=\underset{R^{\beta} \in \{R^0,\,R^1,\,...,\,R^K\}}{max} \left \{\mathcal{TP}_{R^{\alpha} \rightarrow R^{\beta}}(W^{\hat{\rho}}_{b+1}) \right \}
\end{align}

\noindent with $\mathcal{TP}_{R^{\alpha} \rightarrow R^{\beta}}(W^{\hat{\rho}}_{b+1})$ the transition probability which takes into consideration the fact that a regime might still be observed or not. This transition is calculated as follows, 

\begin{align}\label{article2-transition probability}
    \mathcal{TP}_{R^{\alpha} \rightarrow R^{\beta}}(W^{\hat{\rho}}_{b+1}) &= \begin{cases}
        *\;\;if \;\;\; 1\leq\alpha,\beta \leq K\\
        \bm{\Theta}^{\alpha,\;\beta}_{b\; ,\; (b+1)} \cdot \left(1- Surv(W^{\hat{\rho}}_{b+1}\,|\,R^{\beta})\right)\\
        *\;\; if \;\;\; 1\leq\alpha \leq K,\; \beta=0\\
        \bm{\Theta}^{\alpha,\;0}_{b\; ,\; (b+1)} \\
        * \;\; if \;\;\; 1\leq\beta \leq K,\; \alpha=0\\
        \bm{\Theta}^{0,\;\beta}_{j\; ,\; (j+1)}  \cdot \left(1 - Surv(W^{\hat{\rho}}_{j+1}\,|\,R^{\beta})\right)
    \end{cases} 
\end{align}

If the most probable transition goes to none of the regimes $R^r$, $1\leq r \leq K$ (i.e., switching to  $\mathcal{N}_{b+1}$), then the forecast will correspond to a non-observation.  

\begin{algorithm}[tbp]
\small{
 \caption{Forecasting.}
 \label{algo:forecasting}
\KwData{$\mathcal{M}, \; \bm{S},\; W^{\hat{\rho}}_{b+1}$ \scriptsize{\tcc*{A mapping grid of size $(K+1)\times b$, set of time series we want to forecast values at window-stamp $W^{\hat{\rho}}_{b+1}$.}}}
    \KwResult{$\widehat{\bm{S}}_{b+1},\; \widehat{\mathcal{M}}$ \scriptsize{\tcc*{Set of predicted values and the updated mapping grid.}}}  
    \Begin{
        - Define a new mapping grid  $\widehat{\mathcal{M}}$ of size $(K+1)\times (b+1)$ where the $(K+1)\times b$ first values are equal to the one in $\mathcal{M}$ and the last column set to empty sets\;
        - $\widehat{\bm{S}}_{b+1} \gets \emptyset$\;
        - Use a regression model as in Eq. (\ref{article2-generate features}) to predict the general transition $\mathcal{Q}_{b,b+1}$\;
        \textbf{1- Update mapping grid:}\\
        \Begin{
            \For{$S^i \in \bm{S}$}{
                - Using Eq. (\ref{article2-trajectory}) get the trajectory $\mathcal{T}r(S^i|W^{\hat{\rho}}_b)$, the historical features $\{F^{i,\rho}_j\}_{j=1}^b$ of $S^i$ and its latest behavior $R^{\alpha}$\;
                - Use Eq. (\ref{article2-generate features}) to predict the feature $F^{i,\rho}_{b+1}$\;
                - Use a regression model as in Eq. (\ref{article2-generate features}) to predict the transition $\Pi(W^{\rho}_b,\,W^{\rho}_{b+1}\,|\,S^i)$\;
                - Based on $\mathcal{Q}_{b,b+1}$ and $\Pi(W^{\rho}_b,\,W^{\rho}_{b+1}\,|\,S^i)$, use Eq. (\ref{article2-sudden transition3}) to get the effective conditional transition $\bm{\Theta}_{{b}\; ,\; (b+1)}(W^{\hat{\rho}}_b,\,W^{\hat{\rho}}_{b+1}\;|\;S^i) = \left(\bm{\Theta}^{\alpha,\,\beta}_{{b}\; ,\; (b+1)}\right)_{\alpha,\; \beta \in \{0,\,1,\,...,\,K\}}$\;
                - $l \gets \underset{\beta \in \{0,1,...,K\}}{argmax}\left(\bm{\Theta}^{\alpha,\,\beta}_{{b}\; ,\; (b+1)}\right)$ \;
                - $\widehat{\mathcal{M}}[l][b+1] \gets \widehat{\mathcal{M}}[l][b+1] \bigcup \{i\}$ \tcc*{\scriptsize{Populate cell $[l][b+1]$ of the new mapping grid.}}
            }
        }
        \textbf{2- Forecasting:}\\
        \For{$S^i\in \bm{S}$}{
            - Using Eq. (\ref{article2-trajectory}) get the new trajectory $\mathcal{T}r(S^i|W^{\hat{\rho}}_{b+1})$, the historical features $\{F^{i,\rho}_j\}_{j=1}^{b+1}$ of $S^i$ and its latest behavior $R^{\alpha}$\;
            - Use Eq. (\ref{article2-Survival relation}) to calculate each regime survival till window-stamp $W^{\rho}_{b+1}$\;
            - Use Eq. (\ref{article2-forecasted values}) to forecast series values $\widehat{S^{\;i,\;\hat{\rho}}_{b+1}}$ at window-stamp $W^{\hat{\rho}}_{b+1}$\;
            - $\widehat{\bm{S}}_{b+1} \gets \widehat{\bm{S}}_{b+1} \bigcup \{\widehat{S^{\;i,\;\hat{\rho}}_{b+1}}\}$\;
        }
    }
}
\end{algorithm}

To forecast regimes at window-stamps later than $W^{\hat{\rho}}_{b+1}$, we repeat the same process. By running Algorithm \ref{algo:forecasting} we can forecast further regimes that will be exhibited by a given series.

\subsection{Complexity analysis}
In this section, we discuss the time complexity of our proposed framework. Here, it is worth recalling that the proposed framework contains several algorithms, some of which, such as the network clustering algorithm, are not unique choices. In addition, the complexity analysis depends much on intermediate results that are very much data-dependant. This being said, in the current implementation, the regimes' identification and feature learning are the main bottlenecks in the computational cost of the proposed model. In fact, for the regimes' identification step, we test a bunch of time intervals for which we may observe repetitive patterns. For each of the time intervals, we build the corresponding time series network as explained in Section \ref{article2-section:network construction}. Thus, if the series are very long, we may have a large number of time series networks to build. For the feature learning step, it all depends on the suitable selected window size and the number of regimes found for that window. The more we have regimes and time intervals, the more we have clusters for which we have to learn features representing the time series behaviors as shown in Fig. \ref{fig:scan process}$(c)$ (the mapping grid).

Let us suppose that we have $m$ timestamps within which we are observing $N$ time series that evolve together. For a given window of size $\rho$ (with $1 \, < \, \rho \, \leq \, b$), we will have a total number of $b \, = \, m/\rho$ subseries networks where each network $G_j$ edge weights are calculated by the dot product of its membership matrix $\bm{\Delta}_j = (\delta^i_{j,\tau})$, $1\leq i\leq \bm{\mathcal{N}}$, $1\leq \tau \leq 10\,\rho$ with its transpose matrix $\bm{\Delta}^*_j$. Say $\eta_j$ ($\eta_j\leq \rho N$)
is the number of non-zeros in $\bm{\Delta}_j$, its computational complexity is given by $\mathcal{O}(\eta_j \cdot N)$. In the worse case (where $\eta_j = \rho N$), for a specific window of size $\rho$, the computational complexity for passing through the overall time series length to build the networks is given by  $Comp_G(m,\, N,\, \rho)$ 
as follows: 

\begin{align}\label{Complexity_1}
    Comp_G(m,\, N,\, \rho) &= \sum_{j=1}^{m/\rho} \mathcal{O}(\rho \cdot N^2 ) \notag \\
    &= \frac{m}{\rho} \cdot\mathcal{O}(\rho \cdot N^2 ) \notag \\
    &= \mathcal{O}(m \cdot N^2 )
\end{align}

To get the suitable window size for which we can better view the significant regimes, for each constructed network $G_j$ at each window-stamp $W_j$, we perform the Infomap graph clustering \cite{rosvall2008maps} which has a linear time complexity of $\mathcal{O}(|E_j|)$ ($|E_j|$ the number of edges of network $G_j$) \cite{yang2016comparative}. In the worse case (that is when the constructed network is always fully connected), the number of edges at all window-stamp will always be $\frac{N\cdot (N-1)}{2}$. Hence, for a specific window of size $\rho$, the computational complexity for passing through the overall time series length to build the networks is given by $Comp_{clust}(m,\rho)$ as follows:

\begin{align}\label{Complexity_2}
    Comp_{clust}(m,\, N,\, \rho) &= \sum_{j=1}^{m/\rho}\mathcal{O}(E_j) \notag \\
    &= \frac{m}{\rho}\cdot \mathcal{O}\left(\frac{N^2}{2}\right) \notag \\
    &= \frac{1}{2\rho}\cdot \mathcal{O}(m\cdot N^2)
\end{align}

Adding the clustering cost $Comp_{clust}$(Eq. (\ref{Complexity_2})) to the network construction cost $Comp_G$ (Eq. (\ref{Complexity_1})), we then obtain the complexity for the regime identification $Comp_R(m,\, N,\, \rho)$ as follows:

\begin{align}\label{Complexity_3}
    Comp_R(m,\, N,\, \rho) &= Comp_G(m,\, N,\, \rho) + Comp_{clust}(m,\, N,\, \rho) \\ \notag
    &= \left(1+\frac{1}{2\rho}\right) \cdot \mathcal{O}(m\cdot N^2)
\end{align}

Regarding the feature learning, say we have exactly $b$ window-stamps and $K$ regimes represented by $K$ respective clusters at each window-stamp. The computational complexity is calculated with respect to the encoder neural network depth and the number of observations per cluster. In the manuscript, as an encoder, we have used a stack convolutional neural network with two hidden layers having 16 and 8 neurons, respectively. The number of neurons in the input layer, however, varies according to the subnetwork under investigation.

\section{Experiments}\label{article2-experiment}
\subsection{Data description}
We evaluated the proposed approach on ten data sets: one comprising synthetic data and nine, real-world data. In the synthetic data set (SyD), the values of series were generated by $5$ functions as follows:

\begin{align}\label{article2-generator}
    Gn(t) &= \sum_{r=1}^5\,a_r\,\eta_r(t)\,fct_r(t)
\end{align}

\noindent where $a_r\in \{0,1\}$ ( $\sum_{r=1}^5 a_r = 1$) is a parameter allowing one regime to be exhibited at a time by the generated series. $\eta_k(t) \in [0,1]$ is a probability function given in such a way that, at each time-stamp, regimes exhibited by series of the data collection should always respect the following constraint $\sum_{r=1}^5 \eta_r(t) = 1$. 
Functions $fct_r(t)$ are defined as follows:

\begin{align}\label{article2-regime_function}
    fct_r &=
    \begin{cases}
        fct_1(t) = cos(\frac{2\pi t}{5}) + cos(\pi (t-3))\\
        fct_2(t) = sin(\frac{\pi t}{2} - 3) - sin(\frac{\pi t}{6})\\
        fct_3(t) = tan(\frac{\pi t}{2} - 3) -\frac{1}{2}cos(\frac{\pi(t-3)}{6}) + cos(\pi (t-13))\\
        fct_4(t) = sin(\frac{\pi t}{2} - 3)\times cos(\frac{\pi(t-3)}{6}) \times cos(\pi (t-13))\\
        fct_5(t) = cos(\frac{3\pi t}{5}) + sin(\frac{2\pi t}{5} - t)
    \end{cases}
\end{align}

\begin{table}[tbp]
    \centering
    \caption{ Data set description.}
    \label{table:data summarized}
    \begin{tabular}{|c|c|c||c|c|c|}
    \hline
      Data & \#Series & Length & Data & \#Series & Length
     \tabularnewline \hline
       SyD  & $450$ & $1,125$ & RDS & $70$ & $2,844$\\
       ELD  & $100$ & $8,784$ & EQD & $461$ & $512$ \\
       SDSS  & $114$ & $5,112$ & EOG & $724$ & $1,250$\\
       GPS & $50$ & $9,873$ & Pig & $312$ & $2,000$\\
       ESR & $495$ & $4,094$ & CCD & $166$ & $4,307$\\
   Dpt\_1 & $309$ & $5,000$ & Dpt\_2 & $162$ & $5,000$
    \tabularnewline \hline
    \end{tabular}
\end{table}

From the EnerNOC website\footnote{\scriptsize{\url{https://open-enernoc-data.s3.amazonaws.com/anon/index.html}}} (EnerNOC GreenButton Data), we collected the customer electricity load (ELD) data for the whole year $2012$. The read period for each customer was set to $1$ hour. From the Smart Dataset for Sustainability (SDSS)\footnote{\scriptsize \url{http://traces.cs.umass.edu/index.php/Smart/Smart}}, we collected apartment electricity consumption  for the period ranging from 2016-01-01 to 2016-07-31 ($213$ days), with a read set to 1 hour. From the UCI archive \footnote{\scriptsize{\url{https://archive.ics.uci.edu/ml/datasets/}}}, we collected the Gesture Phase Segmentation (GPS) and the Epileptic Seizure Recognition data (ESR).
From the UCR Time Series Classification Archive\footnote{\scriptsize{\url{https://www.cs.ucr.edu/\~eamonn/time\_series\_data/}}}, we collected real world data sets from
the rock data set (RDS), the earthquake data (EQD), electrooculography signal (EOG), pig data (Pig) and chlorine concentration data (CCD). From the Stanford Large Network Dataset Collection\footnote{\scriptsize{\url{https://snap.stanford.edu/data/email-Eu-core-temporal.html}}}, we collected the number of message(s) sent between colleagues of two different departments (Dpt\_1 and Dpt\_2) of an European research institution on an hourly basis. Each of the department is taken as data set. As summarized in Table \ref{table:data summarized}, in these data sets, the number of series varies from $50$ to $750$, while the length of the series varies from $512$ to $9,873$. The selected data sets have different characteristics, which allows us to conduct an objective empirical study to illustrate the suitability of our proposed method to discover  regime shift patterns from co-evolving time series and to make use of these patterns to improve time series forecasting.

Note that, to have more cases with missing series values, for all data sets, we randomly selected series and time intervals from which values would be deleted. To do this, we used the following procedure: Start by randomly selecting a number of series as follows: 

$$Delete(\bm{S}) = \{S^{i}\; |\; i\in randint(1,\,N,\,\;randint(1,\;\frac{N}{30})) \}$$
\noindent where $randint(a, b, c)$ is a uniform random function that generates $c$ integers within the interval $[a,\;b]$, whereas $randint(a,\;b)$ generates one integer within the interval $[a,\;b]$. Then, randomly select a time interval within which series values in $Delete(\bm{S})$ will be deleted, as follows:

\begin{align*}
    Delete(m) &= \left\{\{(t_j,\,t_{j+\rho})\}\,|\, \rho \in randint(\frac{m}{30},\,\frac{m}{10}),\, j \in randint(1,\,m-\rho) \right\}
\end{align*}
\noindent with $m$ the length of series in $\bm{S}$.

In what follows, we start by illustrating the various important steps which are required for predicting the series values at further time points. Later on, we present the forecasting results and assess the accuracy of our model compared to some well-known state-of-the-art time series forecasting methods.

\begin{figure}[tbp]
    \centering
    \subfloat{\includegraphics[width=1.97in]{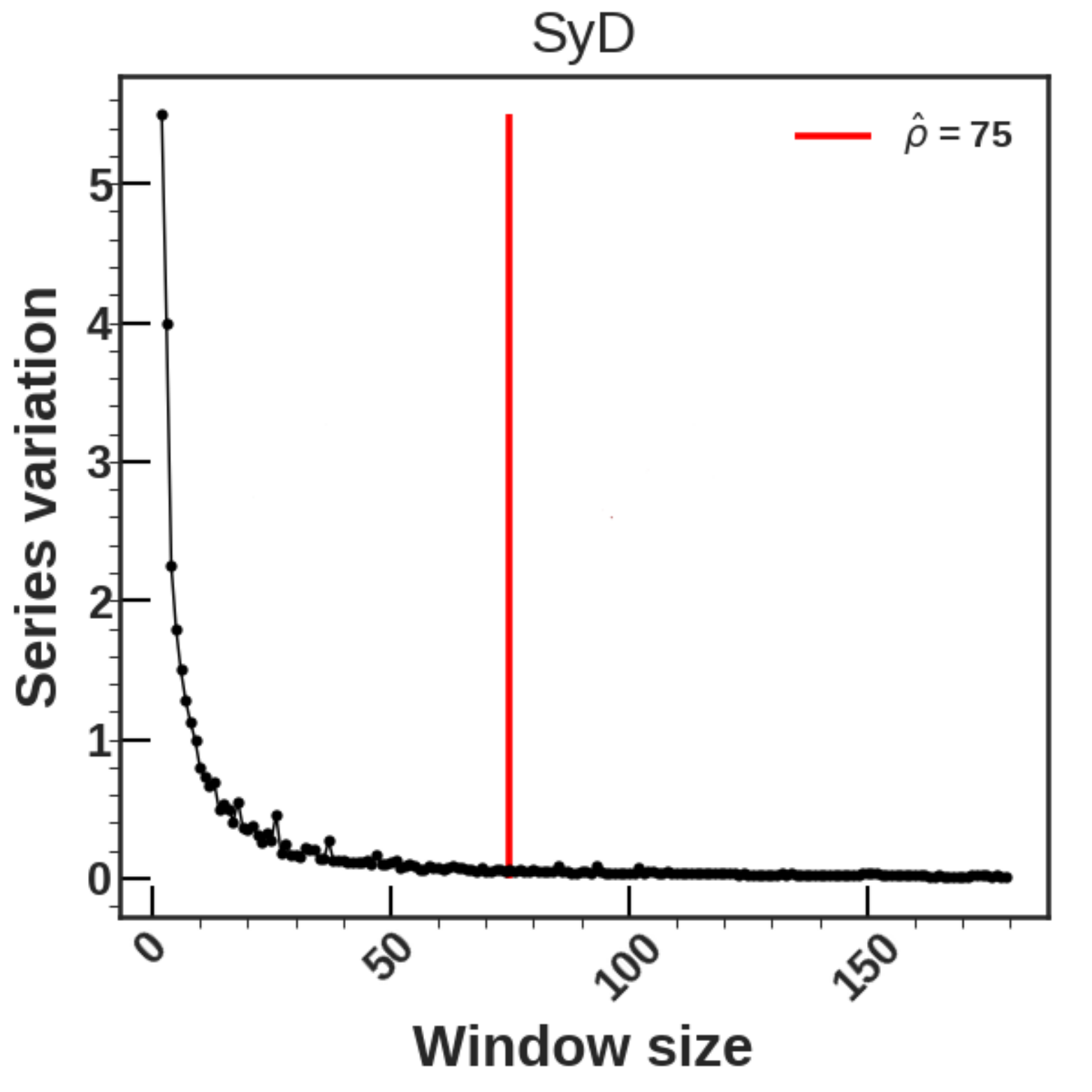}}\hspace*{-1mm}
    \subfloat{\includegraphics[width=1.97in]{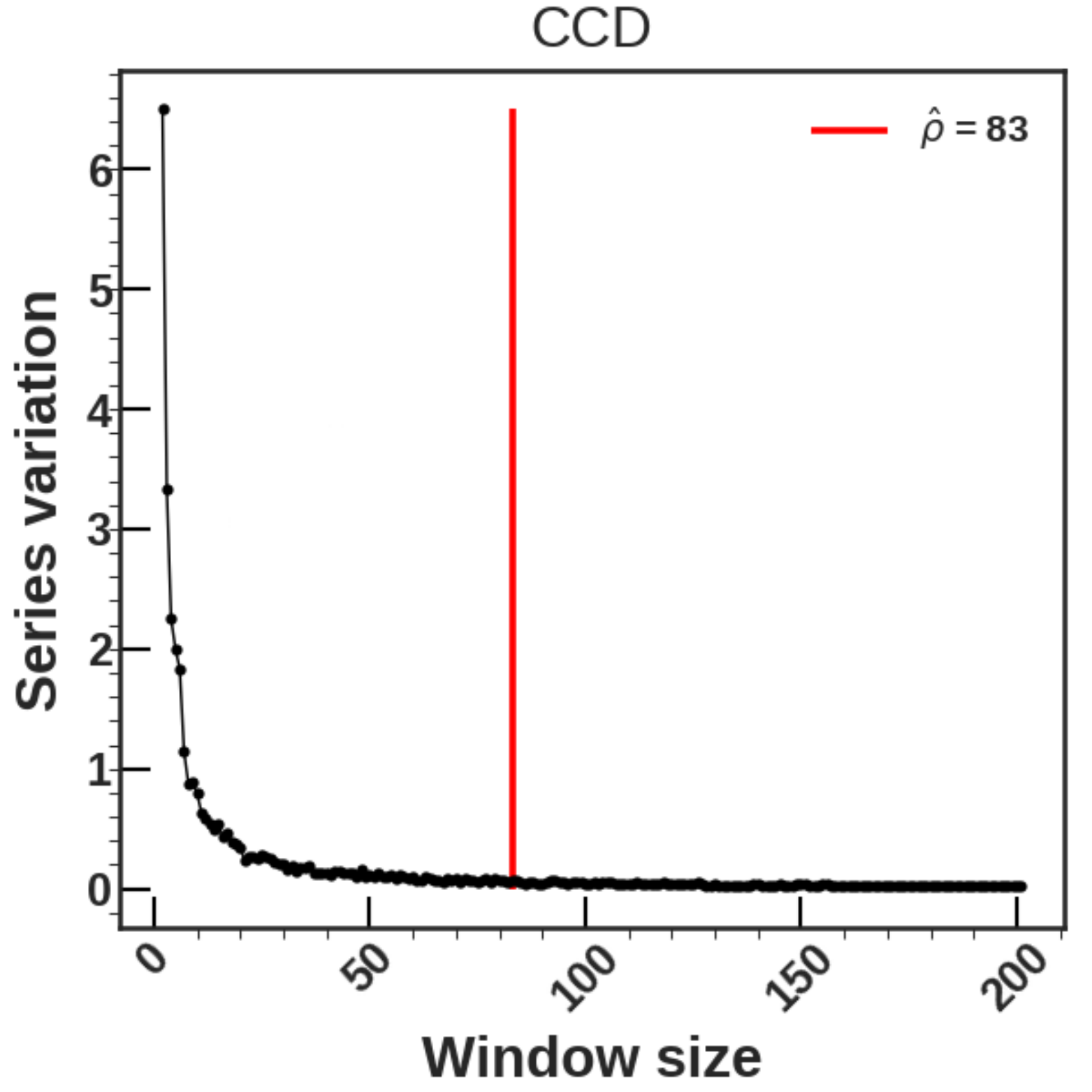}}\vspace{-2mm}
    \caption{ Example of \textit{regime-density-per-window-size} in SyD and CCD data sets. The vertical red line gives the optimal window size in each data set.}
    \label{fig:series variation}
\end{figure}

\subsection{Data scanning}
We identified the optimal window size for each of the ten data sets considered in the experiments. Specifically, we applied Algorithm \ref{algo:down drill} to build  and split the networks related to subseries at each instance of a given window, and to identify the optimal window size and the groups of subseries exhibiting the different hidden regimes.

\subsubsection{Window's size \& regime identification}
Based on the relation given in Eq. (\ref{article2-series_dynamic}), we estimate how likely co-evolving series in each of our data sets are to vary. For the purpose of illustration, in Fig. \ref{fig:series variation}, we have the \textit{regime-density-per-window-size} scores in the synthetic data set (SyD) and the Chlorine Concentration data set (CCD).
As depicted, we see that the density scores in all cases drops rapidly when the size of the window is small. As the size of the window gets larger, the density score tends to converge to a constant value. The values associated with the vertical red lines in Fig. \ref{fig:series variation} show the optimal window sizes for which we can best track regimes over time. Note that the obtained window size varies depending on the data.

\begin{figure}[tbp]
    \centering
    \subfloat{\includegraphics[width=4.7in]{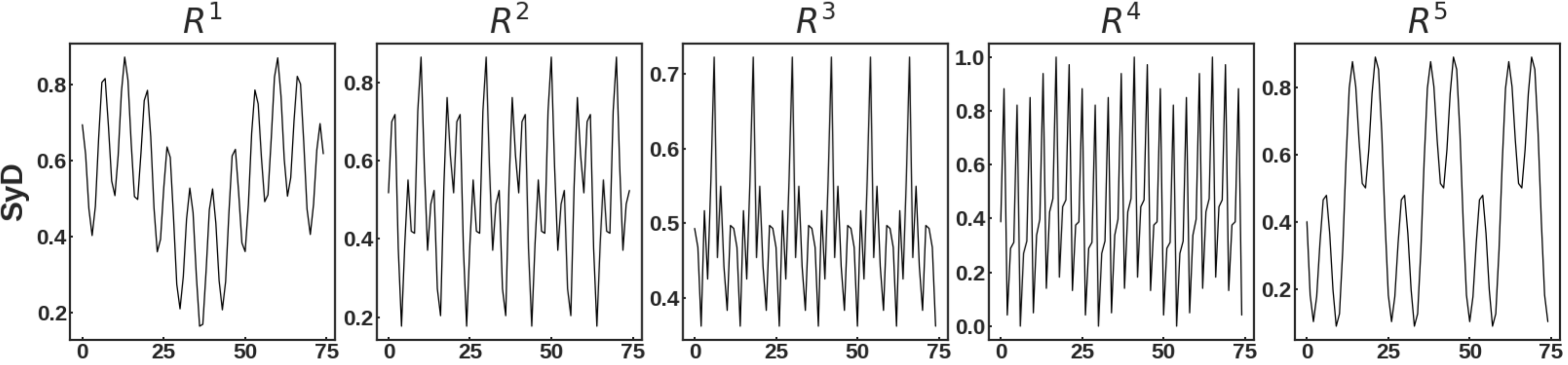}}\\
    \subfloat{\includegraphics[width=4.7in]{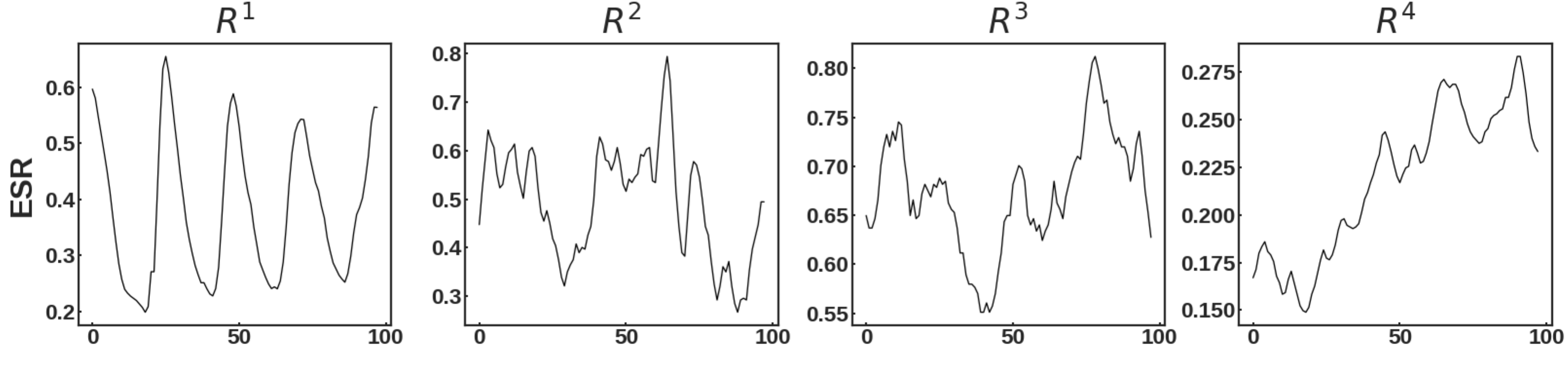}}\\
    \subfloat{\includegraphics[width=4.7in]{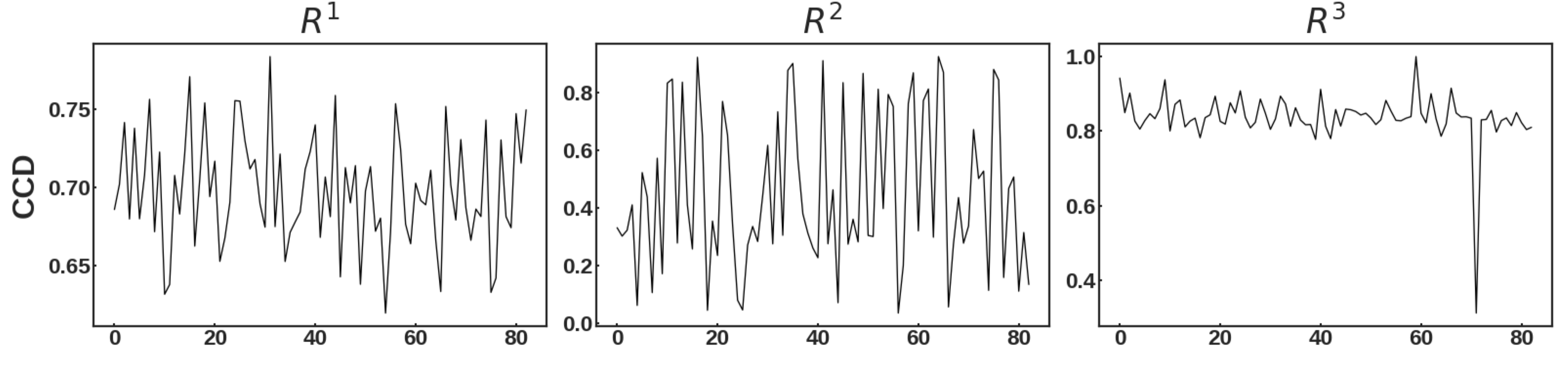}}
    \caption{ Example of identified regimes in SyD, ESR and CCD.}
    \label{fig:exhibited regimes}
\end{figure}

Using the optimal window sizes $\hat{\rho}$, we then look for the distinct regimes exhibited by series. In Fig. \ref{fig:exhibited regimes}, for each row, we have the distinct regimes exhibited by co-evolving series in the SyD, ESR and CCD data sets respectively. As can be seen, when running Algorithm \ref{algo:down drill}, we found that the number of regimes $K$ in the SyD, ESR and CCD data sets is, respectively, $5$, $4$ and $3$. Note that for the specific case of SyD the number of regimes $K = 5$ again corresponds to the number of functions  used in Eq. (\ref{article2-regime_function})  to generate the synthetic data. For real cases, we do not have the ground truth for validating the obtained regimes. However, since the forecasting quality depends on the identified regimes, we will be able to assess whether the identified regimes are the right ones. Table \ref{table:resume} provides the optimal window size and the number of regimes for each of the tested datasets.

\begin{table}[tbp]
    \centering
    \caption{ Window Sizes and number of exhibited regimes.}
    \label{table:resume}
    \small{
    \begin{tabular}{|c|c|c|c|c|c|c|c|}
    \hline
      Data & $\hat{\rho}$ & \#Regimes & Data & $\hat{\rho}$ & \#Regimes
     \tabularnewline \hline
       SyD  & $75$ & $5$ & RDS & $109$ &  $5$\\
       ELD  & $94$ & $2$ & EQD & $53$ & $3$\\
       SDSS  & $75$ & $2$ & EOG & $181$ & $5$\\
       GPS  & $103$ & $8$ & Pig & $75$ &  $4$\\
       ESR  & $98$ & $4$ & CCD & $83$ & $3$ \\
       Dept\_1 & $72$ & $6$ & Dept\_2 & $48$ & $8$
    \tabularnewline \hline
    \end{tabular}
    }
\end{table}

\subsection{Mapping Grid}

\subsubsection{Grid construction}
\begin{figure}[tbp]
    \centering
    \subfloat{\includegraphics[width=3.75in]{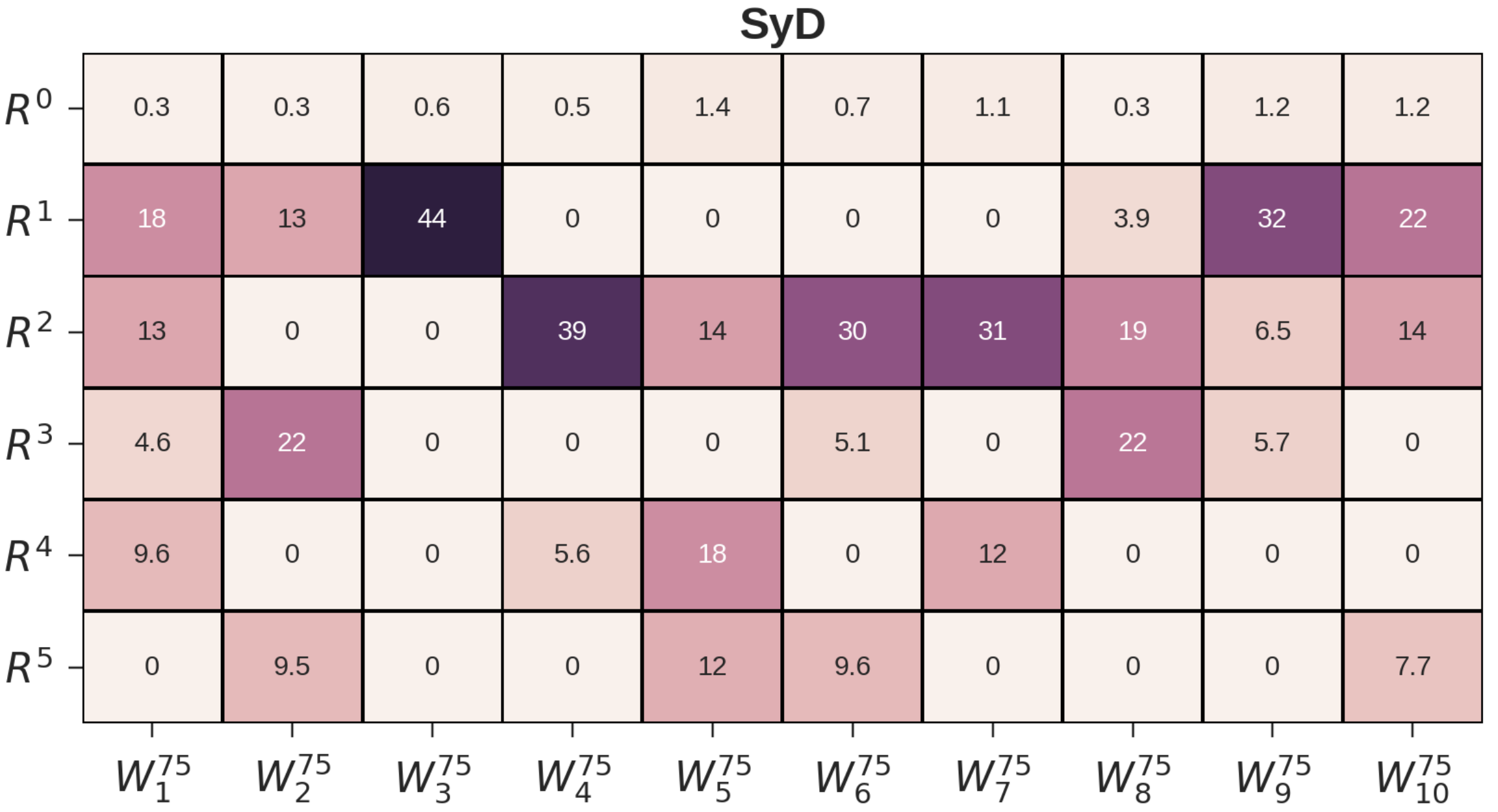}}\vspace{-5pt}
    \subfloat{\includegraphics[width=4.in]{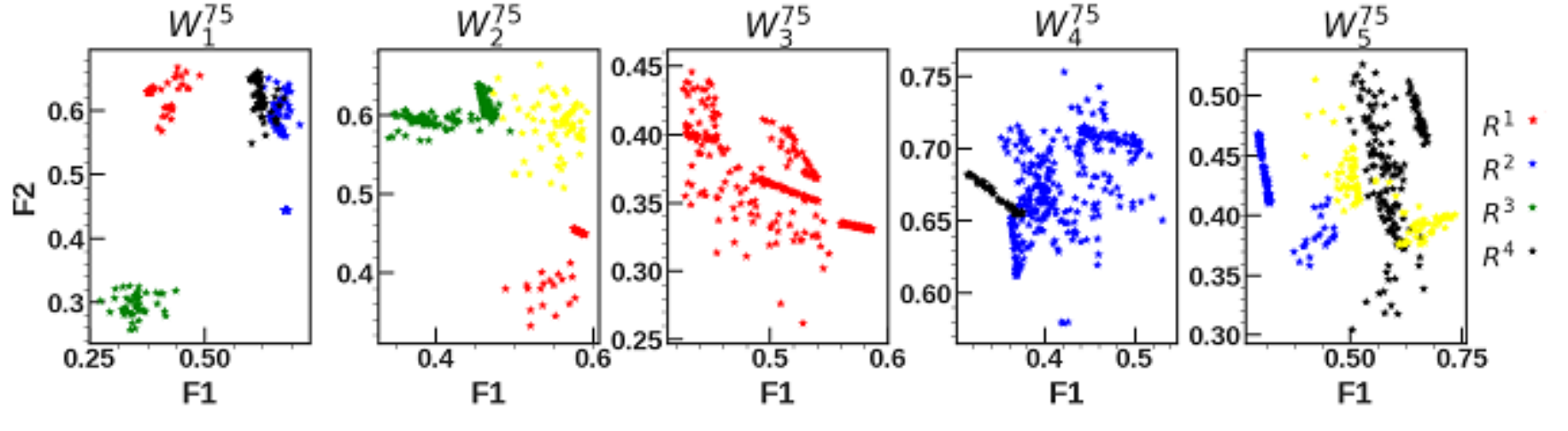}}\vspace{-5pt}
    \subfloat{\includegraphics[width=3.75in]{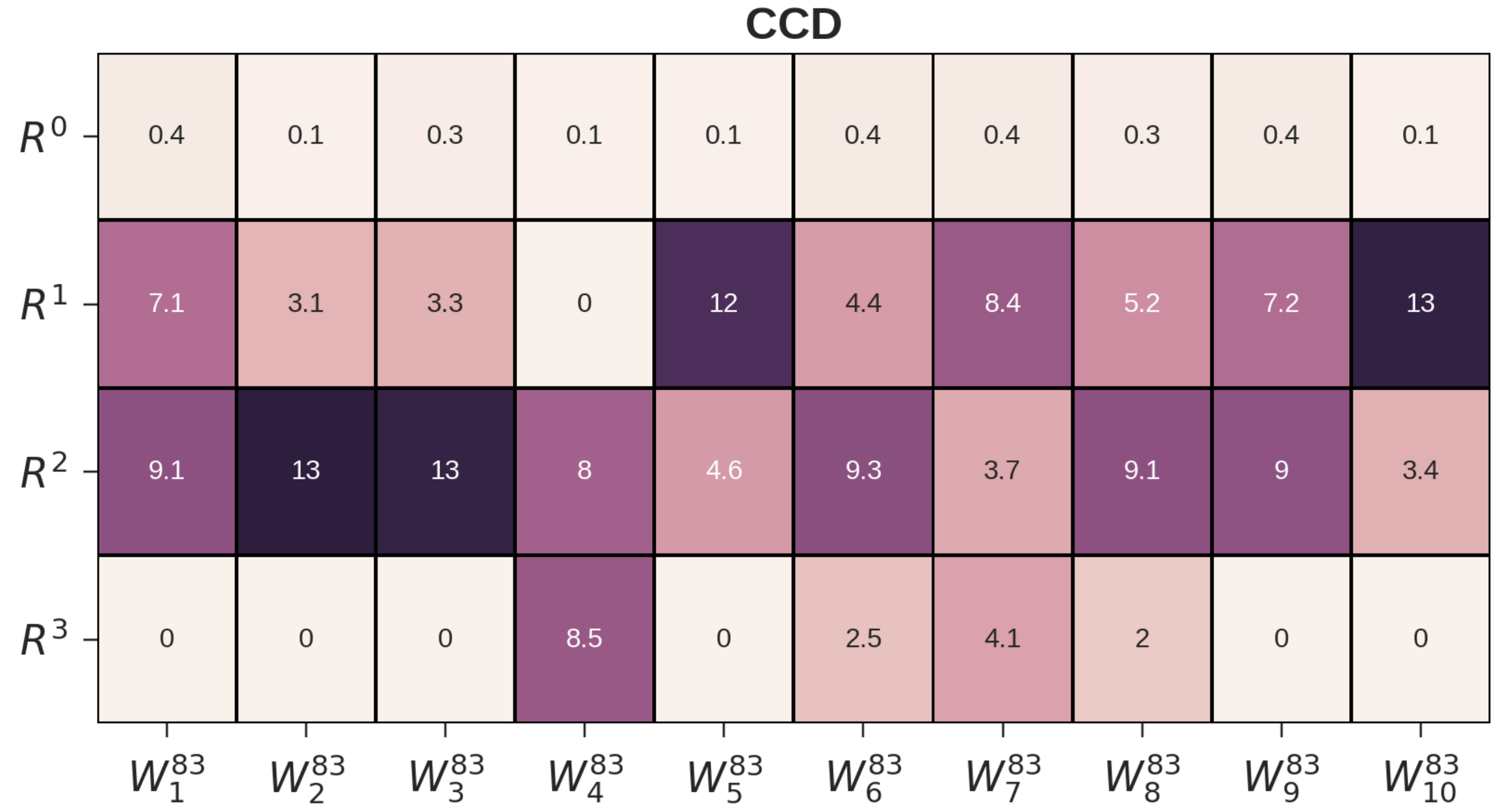}}\vspace{-5pt}
    \subfloat{\includegraphics[width=4.in]{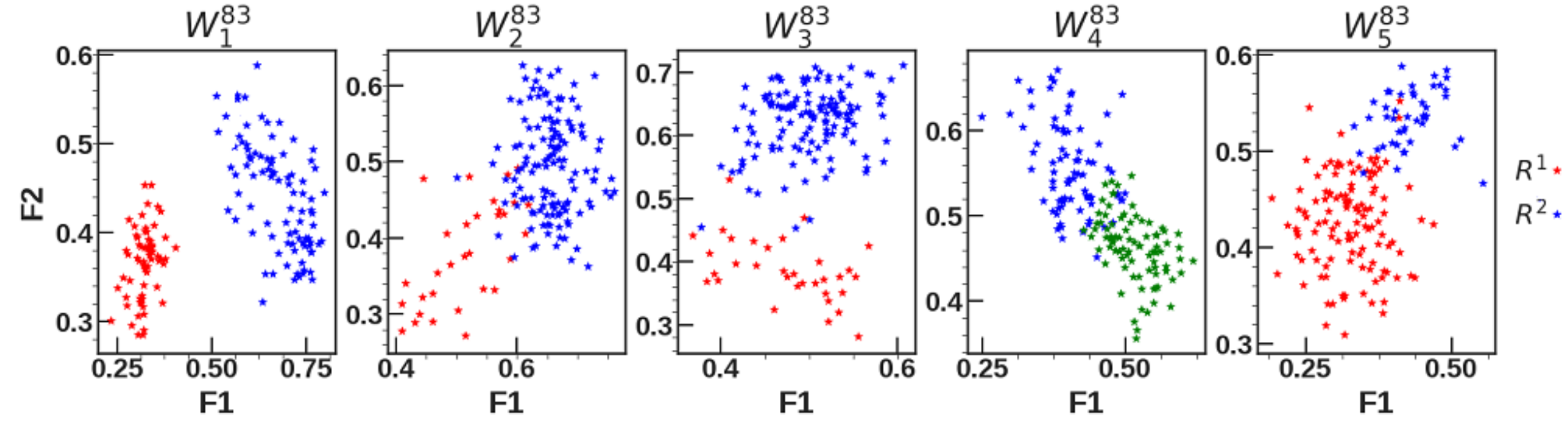}}
    \caption{ Example of mapping grids in SyD and CCD at the $10$ first window-stamps. Below each mapping grid, we have the embedding in a 2-D projection at the $5$ first window-stamps.}
    \label{fig:mapping grids}
\end{figure}

With the discovered regimes in Fig. \ref{fig:exhibited regimes}, we now identify series that contain each of these regimes at the different window instances that constitute our mapping grid. From each cell of the mapping grid, we also learn features that encompass the subseries relationships, which helps in understanding regime lifespans. For the purpose of illustration, in Fig. \ref{fig:mapping grids} we have two examples of mapping grids in SyD and CCD with their corresponding embeddings that stand for the learned features. The heatmaps show only the first 10 window-stamps in each case. In these heatmaps, the darker the cell, the greater the number of series exhibiting the regime. It can be seen that the colors of the heatmaps corresponding to the data sets depicted in Fig. \ref{fig:mapping grids} change at different window-stamps. This shows that there are indeed series that contain different regimes at different window-stamps. It can be seen that, at some rows of our grids (e.g., $R^2$ in SyD and $R^1$ in CCD respectively), the color gradually becomes dark when reading from left to right, showing that the corresponding regime is more and more observed in most of the series as time advances. Similarly, for some rows of our grids (e.g., $R^1$ in SyD and $R^2$ in CCD respectively), the color becomes light when reading from left to right, showing that the corresponding regime is less and less exhibited by series as time advances.

\subsubsection{Features learning}
To learn features from each cell of the mapping grid, we make use of a graph autoencoder. Specifically, knowing that each cell of the mapping grid is populated by a subnetwork $G^r_j$, we use this subnetwork as the input to our graph autoencoder to generate an embedding $\{F^{\;i,\,\hat{\rho}}_j\}$ that captures both the topological structure of $G^r_j$ and its content. The obtained embedding is then used as the features of the mapping grid cell. It is worth noting that the embedding of each node of the subnetwork $G^r_j$ corresponds to an $8-dimensional$ vector. In our architecture, we use as encoder a stack convolutional neural network with two hidden layers having $16$ and $8$ filters respectively.

Again, as an example, we show, in Fig. \ref{fig:mapping grids} and below each of the mapping grids, the corresponding 2-D feature projections from various regimes at five consecutive window-stamps of series in SyD and CCD. In these 2-D plots, it can be seen that the embedded space is characterized by the presence of fairly distinguishable structures that allow discriminating between the different regimes. Moreover, from the embedded space, we can also observe the non presence
of some regimes. For example, regime $R^1$ (tagged in red) in CCD and regime $R^5$ (tagged in yellow) in SyD are not observed in the embedded space at the first window-stamp.

Having constructed our mapping grids, we can now utilize the embedded space as our learned features at each window-stamp to better understand the time duration for which a regime may or may not persist. In the following subsection, we illustrate the regime lifespans.

\subsection{Regime survival}
Recall that, to study regime lifespans, we make use of the Cox regression model for which we need to identify the parameters. In this paper, we have devised our Cox regression (Eq. (\ref{article2-Risk on event})) in such a way that it can integrate time-dependent features as well as time-dependent parameters. From what we have presented so far, we can effectively say that the learned features from our mapping grids do change with time. This is why we believe that their importance may change as well from one time to another.

\begin{figure}[tbp]
    \centering
    \includegraphics[width=4.65in]{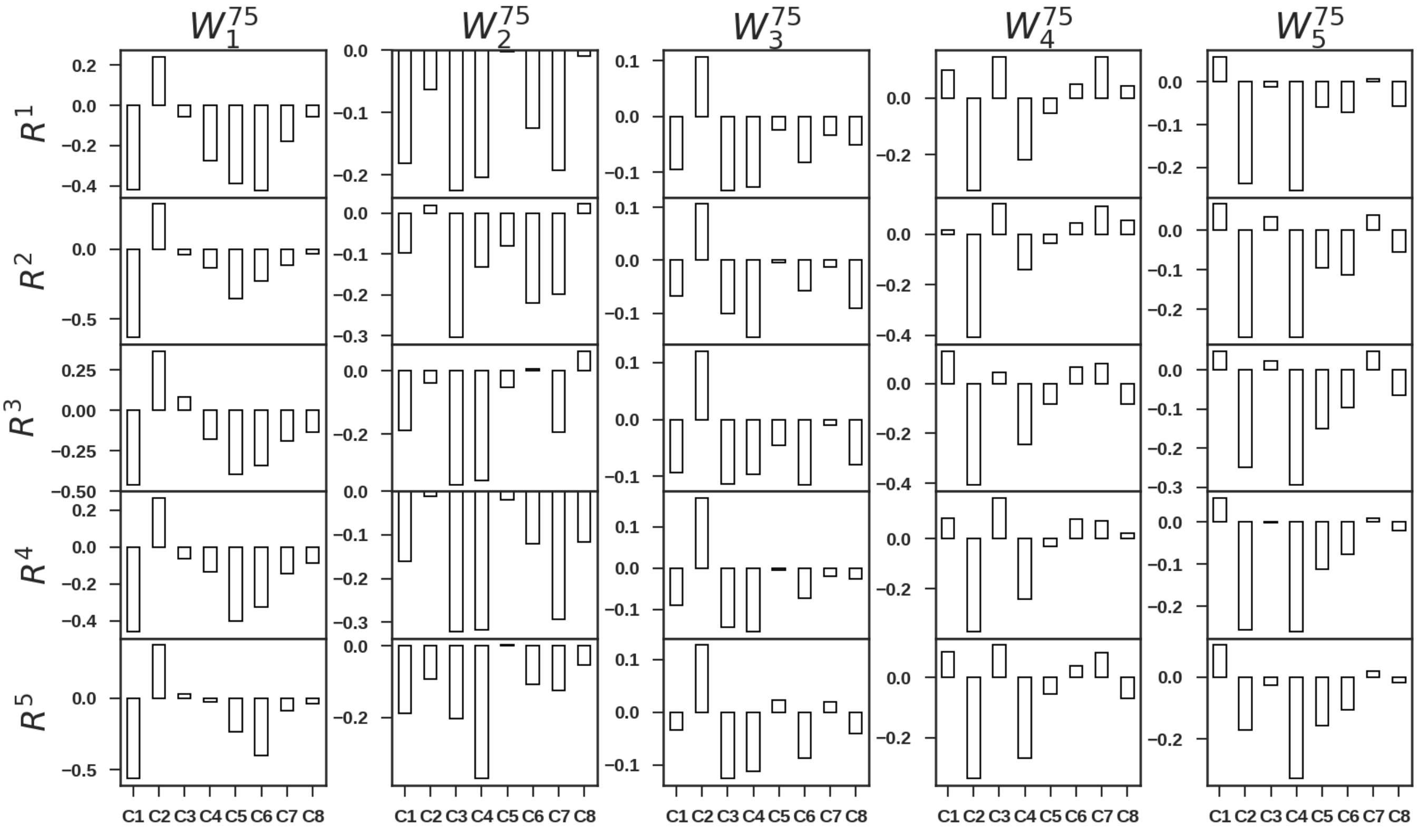}
    \caption{ Cox coefficients at the five first window-stamps in SyD data sets. The variation observed at each window-stamp for each regime demonstrates that the importance of each feature changes with time. Moreover it also shows that the number of series exhibiting each of the regimes changes as well.}\vspace{-3mm}
    \label{fig:cox parameters}
\end{figure}

To demonstrate this, Fig. \ref{fig:cox parameters} presents an example of the extracted Cox parameters over five consecutive window-stamps for regimes identified in SyD. As can be seen, the bar plots change with time, showing that the role, or the importance, of each feature changes over time. 

Based on the Cox parameters and the feature vectors corresponding to series in different groups of subseries at different window-stamps, we can thus evaluate the lifespan probability of each regime. Recall that, in all our experiments, for each regime $R^r$, we make use of the Gamma distribution as the baseline hazard function ($\gamma^r()$ in Eq. (\ref{article2-Risk on event})). The gamma distribution parameters are estimated using the experimental distribution corresponding to the number of series not exhibiting regime $R^r$ over time. 

\begin{figure}[tbp]
    \centering
    \subfloat{\includegraphics[width=1.52in]{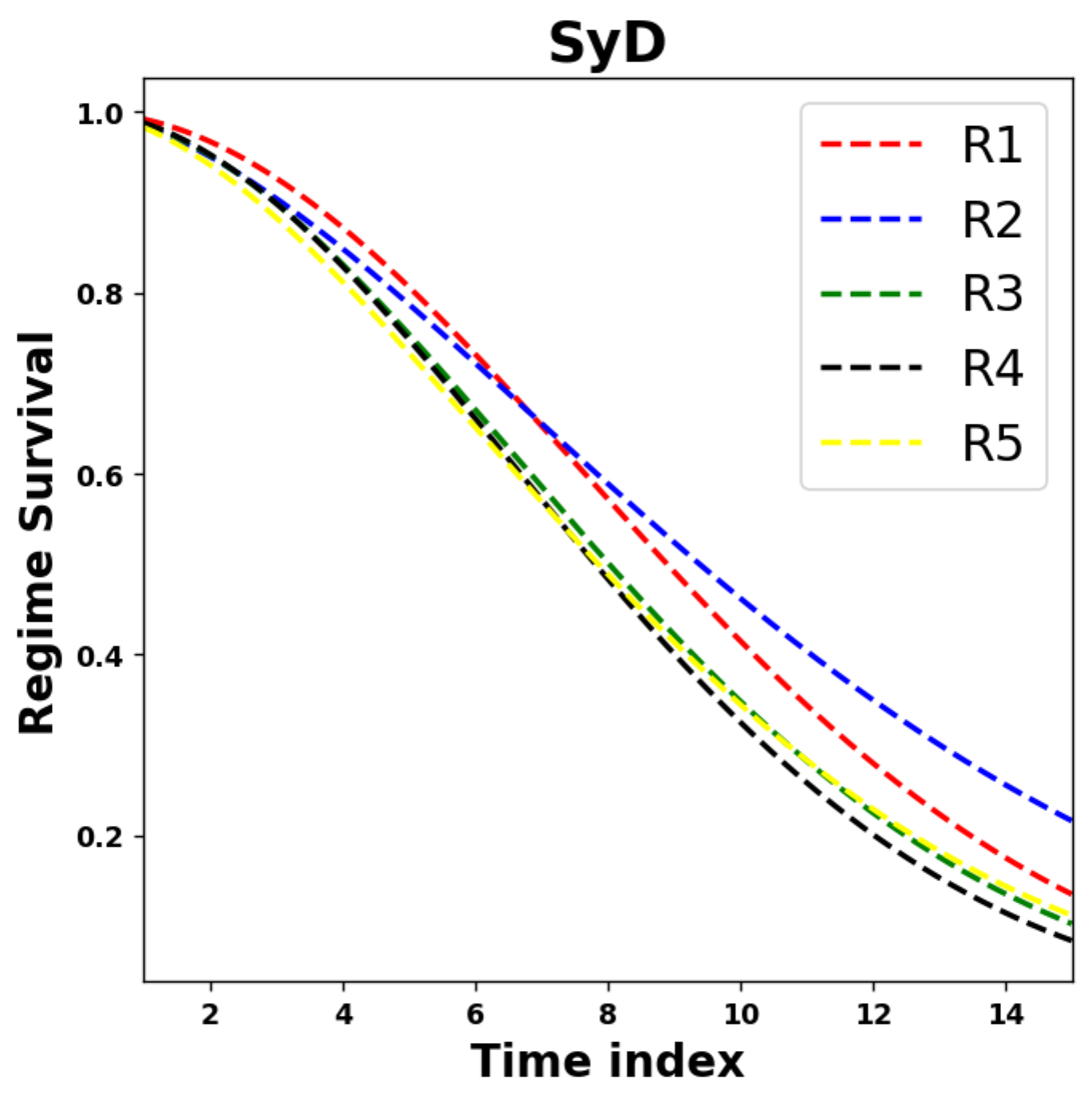}}
    \subfloat{\includegraphics[width=1.52in]{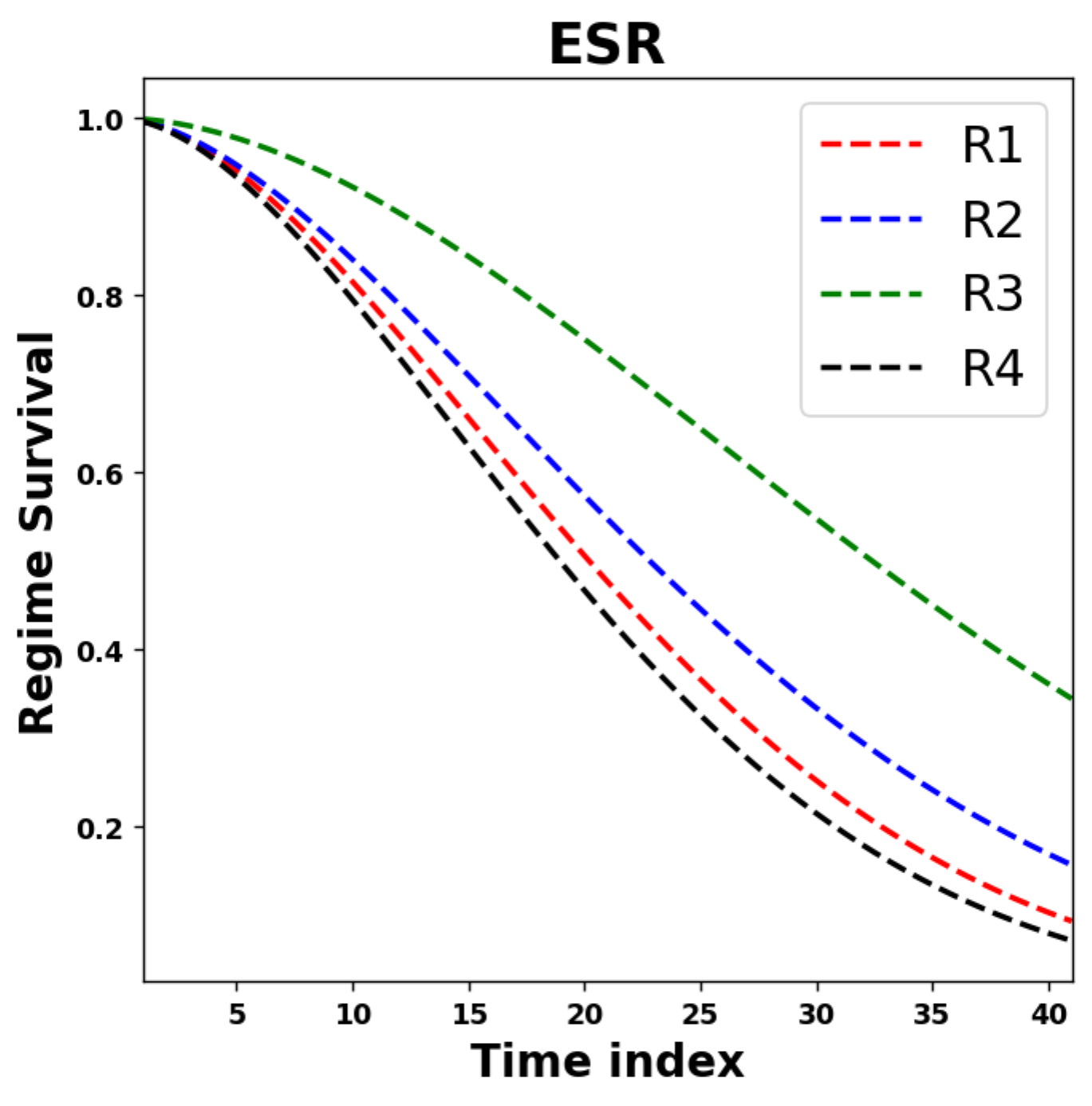}}
    \subfloat{\includegraphics[width=1.52in]{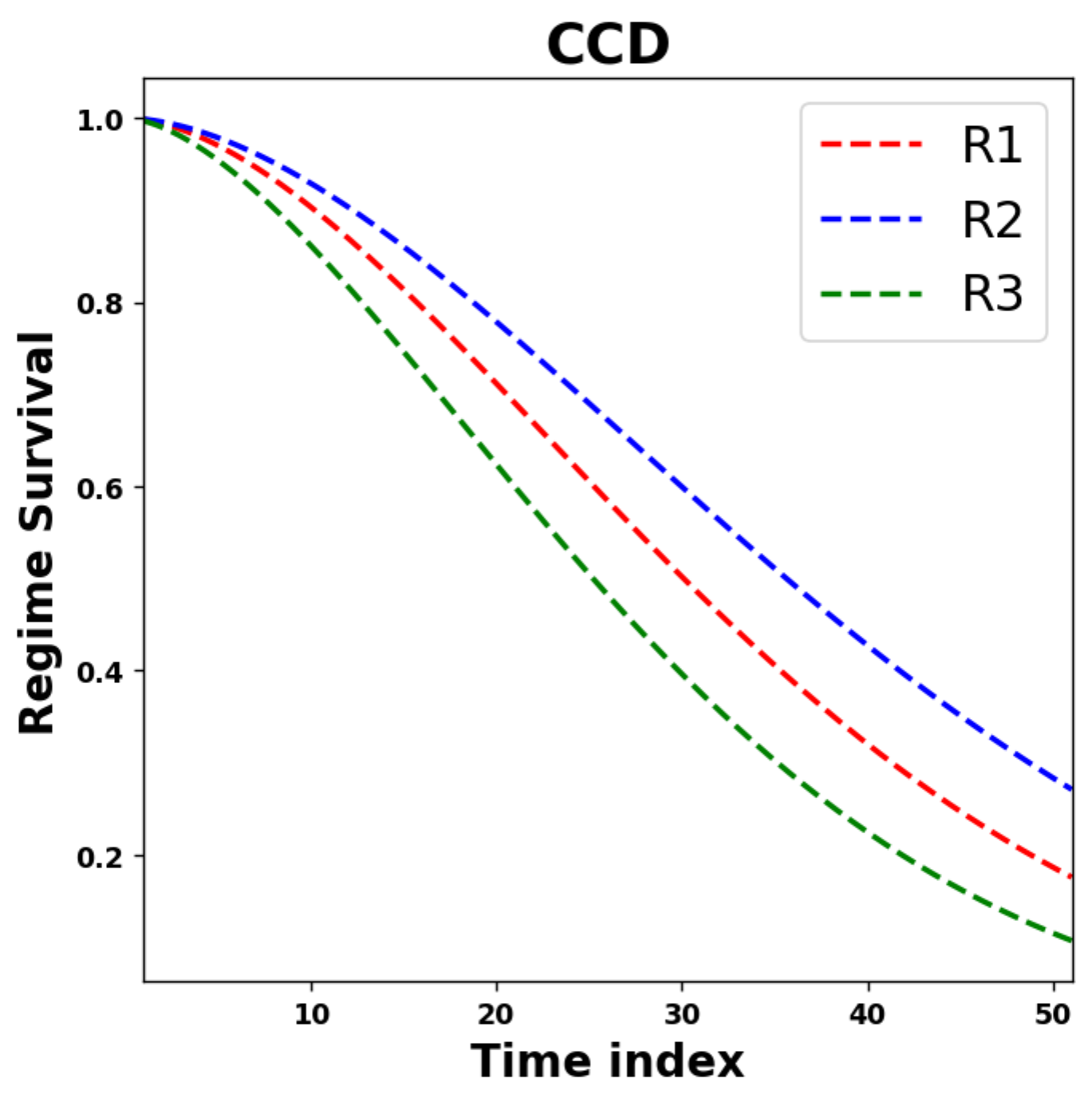}}
    \caption{ Examples of regime lifespans in SyD, ESR and CCD.}
    \label{fig:regime lines}
    \vspace{-10pt}
\end{figure}

Using the baseline function, the learned features and the Cox parameters, we now apply Eq. (\ref{article2-Survival relation}) to obtain the probability of a regime no longer being exhibited over time (the regime lifespan). In Fig. \ref{fig:regime lines} we have examples of regime lifespans in SyD, ESR and CCD. As depicted, we note that, as time evolves, the probability of a regime not being exhibited decreases quickly after a certain amount of time in all cases. For some regimes, this lifespan probability drops very fast compared to others that remain stable after a certain time. In the ESR data set, for instance, the plot shows that the probability of regimes $R^1$ $R^2$ and $R^4$ to be observed drops faster than $R^3$. This means that, as time evolves, regime $R^3$ tends to persist in ESR compared to other regimes. The same interpretation is done in the SyD and CCD data sets, where we can note that regime $R^2$ (in SyD and CCD) has a higher probability to be latter observed compared to other regimes.

\begin{figure}[tbp]
    \centering
    \subfloat{\includegraphics[width=4.in]{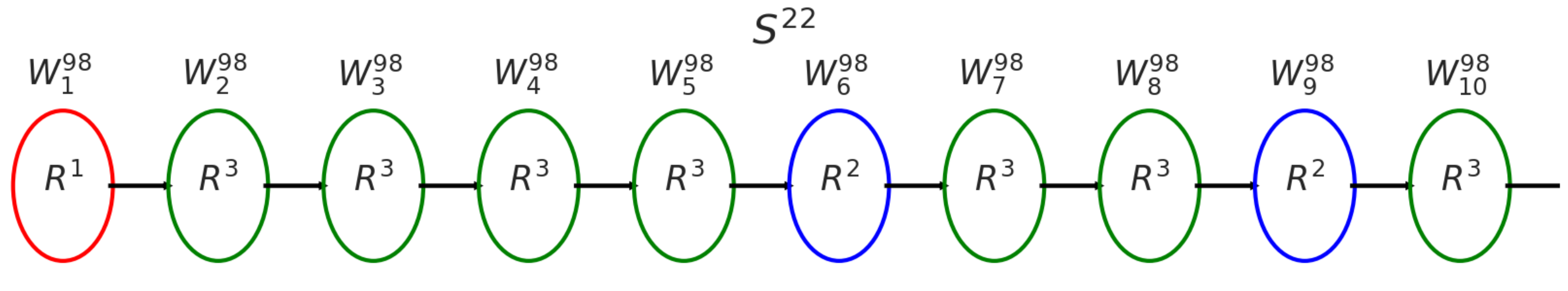}}\\
    \subfloat{\includegraphics[width=1.36in]{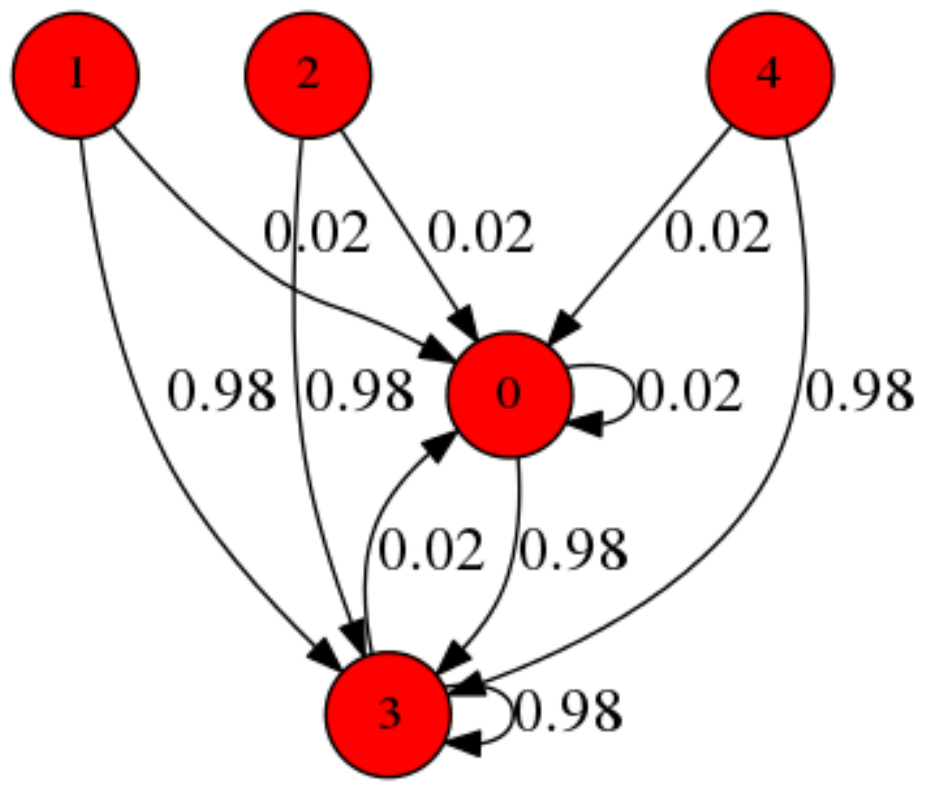}}
    \subfloat{\includegraphics[width=1.36in]{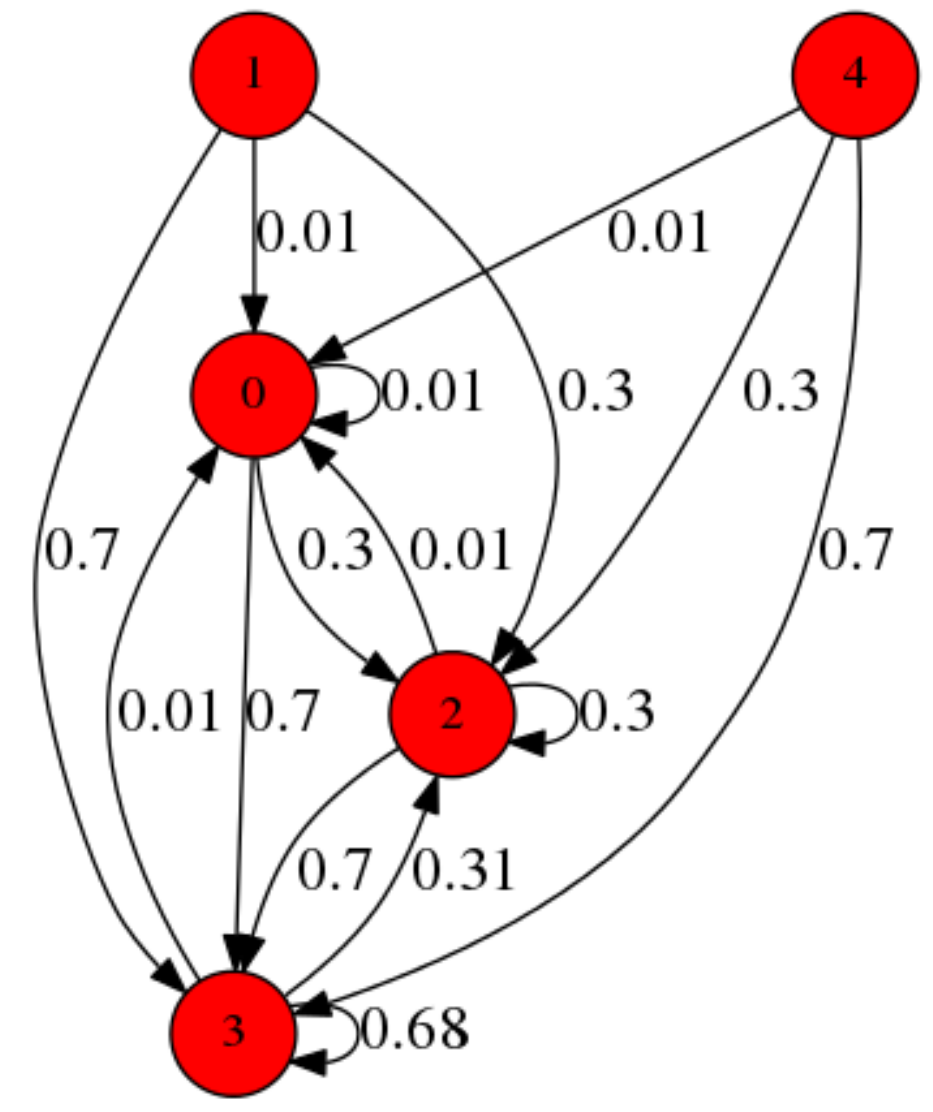}}
    \subfloat{\includegraphics[width=1.36in]{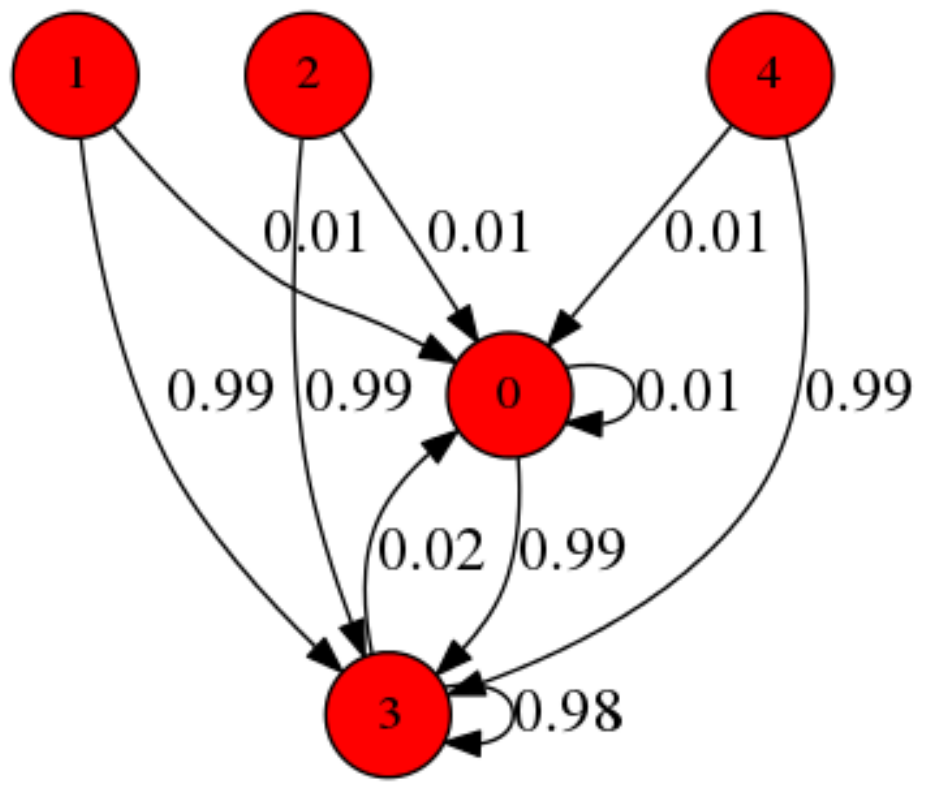}}
   \\
    \subfloat{\includegraphics[width=4.in]{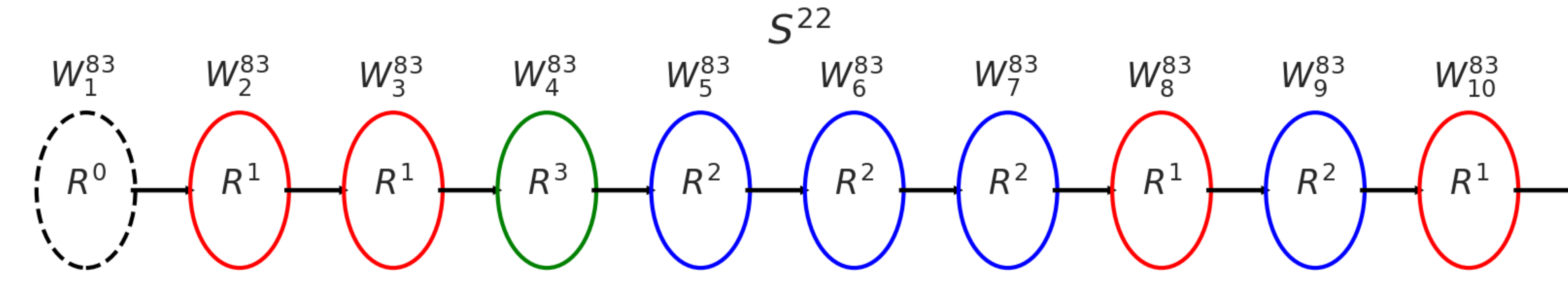}}\\
    \subfloat{\includegraphics[width=1.12in]{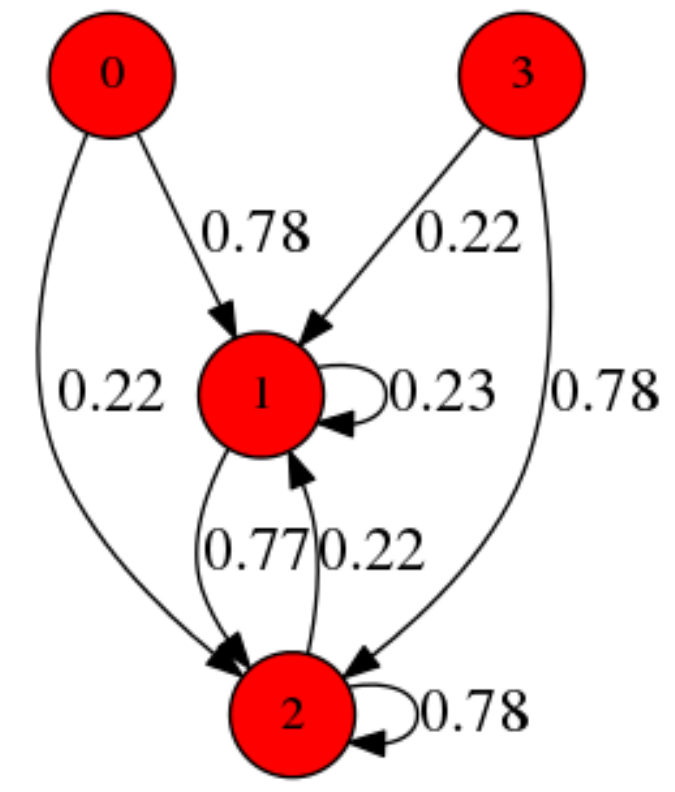}}
    \subfloat{\includegraphics[width=1.5in]{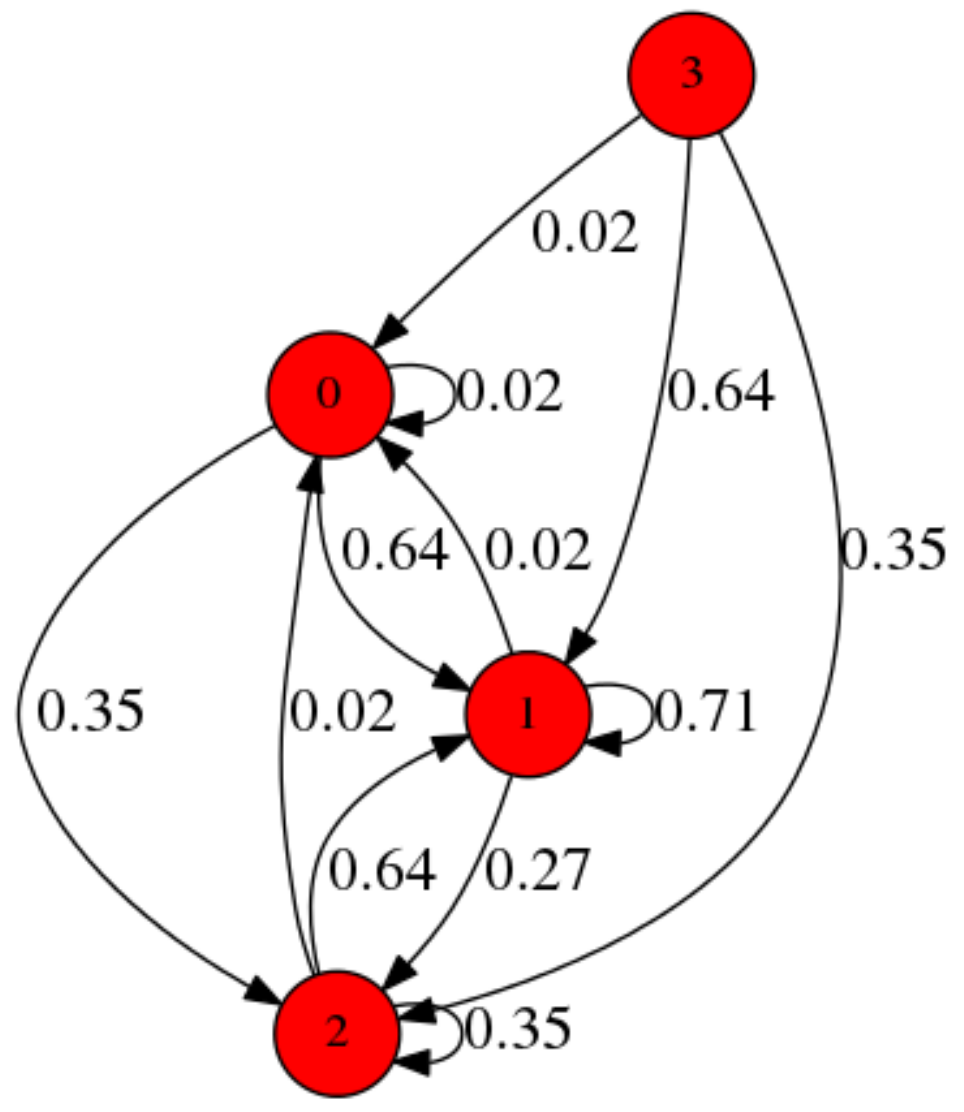}}
    \subfloat{\includegraphics[width=1.12in]{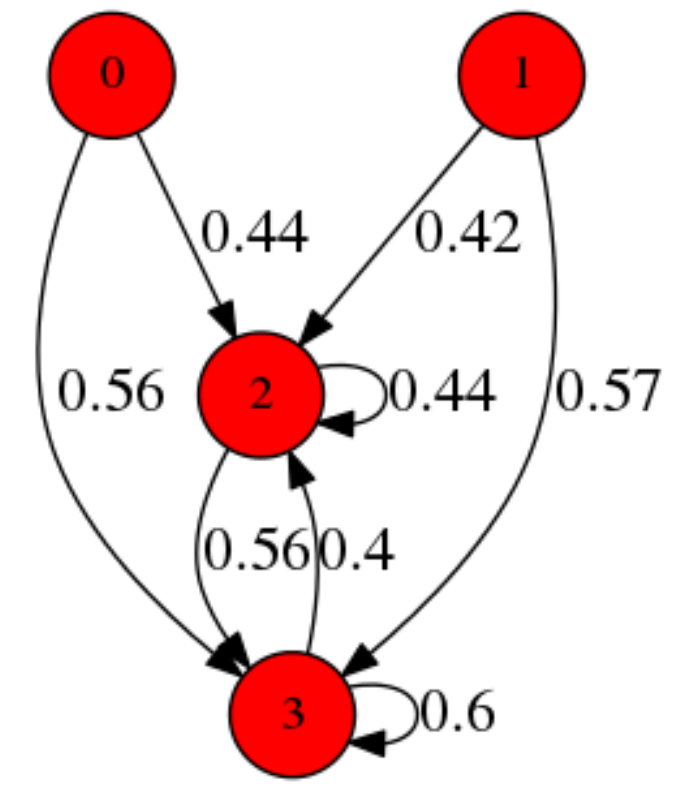}}
    \caption{ Trajectory followed by series $S^{22}$ in the ESR and CCD data sets at the first ten window-stamps ($W^{98}_1$ $-$ $W^{98}_{10}$ in ESR and $W^{83}_1$ $-$ $W^{83}_{10}$ in CCD). Graphs below each trajectory depict the first three conditional transition matrices ($\bm{\Theta}_{1 , 2}$ $-$ $\bm{\Theta}_{3 , 4}$) extracted from each of the trajectories of $S^{22}$.}
    \label{fig:series trajectory}
\end{figure}

\subsection{Forecasting}
For each data set, we consider all window-stamps to be known except the last four, for which we want to predict the series values. For the known window-stamps, we start by extracting series trajectories (with the corresponding features) and their respective conditional transitions. From the extracted information, we train a generative model in such a way to make it suitable for predicting features and transition probabilities at more distant window-stamps.

\subsubsection{Series trajectory \& regime shift}
To track series behavior, we utilize our \textit{mapping grids}. Fig. \ref{fig:series trajectory} presents examples of time series trajectories from the first window-stamp till the tenth window-stamp in the ESR and CCD data sets. The network below each trajectory corresponds to the transition matrix at the first four consecutive window-stamps. Each circle of the trajectory indicates the regime exhibited at a specific window-stamp. Here, from the sequence of circles, it can be observed that the exhibited regimes are non-contiguous over time by looking at the transition matrices for each of the series. Within two consecutive window-stamps, it can be seen that the transition probability between regimes changes with time. Moreover, the different transition matrices explain the behavioral sequence in each case. 

More precisely, let us take for instance the sequence $S^{22}$ in the CCD data set. From window-stamp $W^{83}_1$ to $W^{83}_2$, we note that series $S^{22}$ has passed from behavior $R^0$ to $R^1$ and that the conditional transition $\bm{\Theta}^{0,1}_{1,2} = 0.78$ is greater than $\bm{\Theta}^{0,2}_{1,2}$. This means that the transition matrix $\bm{\Theta}_{1,2}$ can well explain the changing behavior that happened from $W^{83}_1$ to $W^{83}_2$. At the next consecutive window-stamps (from $W^{83}_2$ to $W^{83}_3$), we note that the behavior remains the same. If we were to use the same transition matrix (the one used at consecutive window-stamps $W^{83}_1$ to $W^{83}_2$, $\bm{\Theta}_{1,2}$), the series should switch to regime $R^2$, which does not correspond to the sequence. Looking instead at the transition matrix $\bm{\Theta}_{2,3}$, it corroborates the fact that the series keeps the same behavior at window-stamp $W^{83}_3$. The same observations are made at subsequent window-stamps in all cases (ESR and CCD).

\begin{figure}[tbp]
    \centering
    \subfloat[\small SyD]{\includegraphics[width=4.65in]{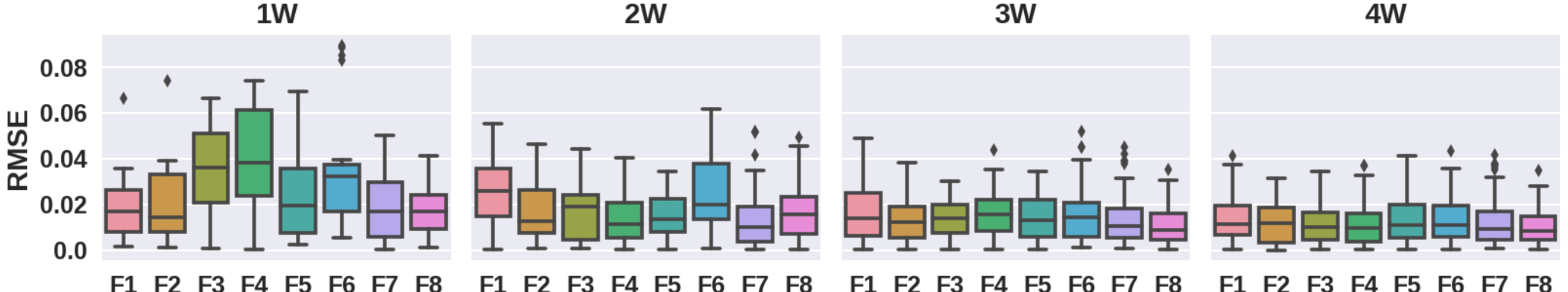}}
    \\
    \subfloat[\small ESR]{\includegraphics[width=4.65in]{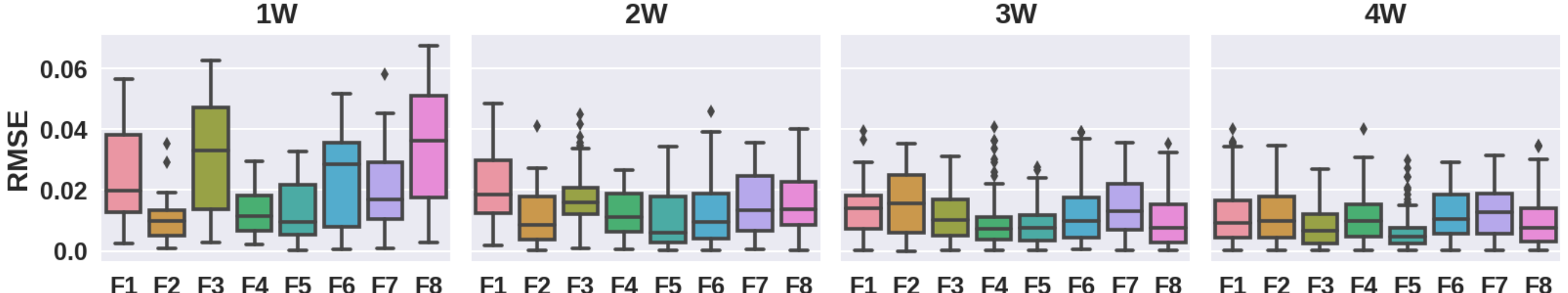}}
    \\
    \subfloat[\small CCD]{\includegraphics[width=4.65in]{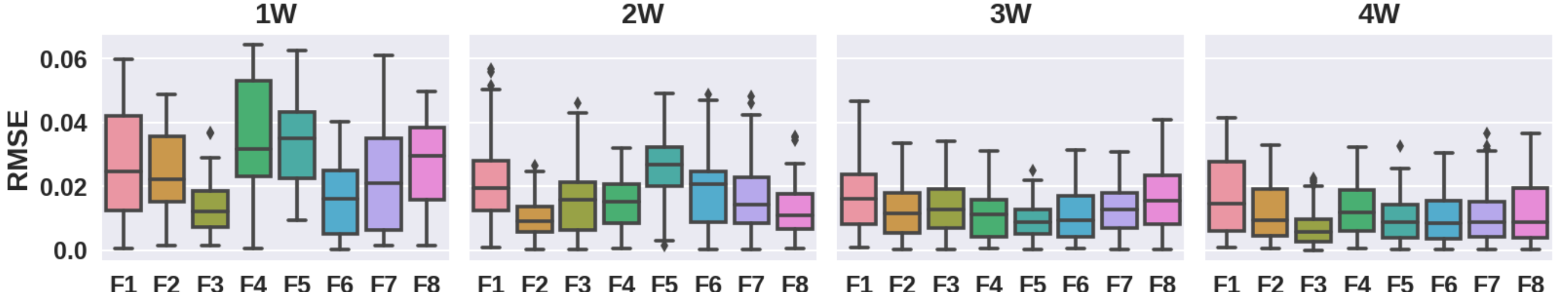}}
    \caption{ Example of error made (RMSE) when predicting features values at 1, 2, 3 and 4 window-stamps ahead in SyD, ESR and CCD.}
    \label{fig:error feature forecasting}
\end{figure}

\begin{figure}[tbp]
    \centering
    \subfloat{\includegraphics[width=1.56in]{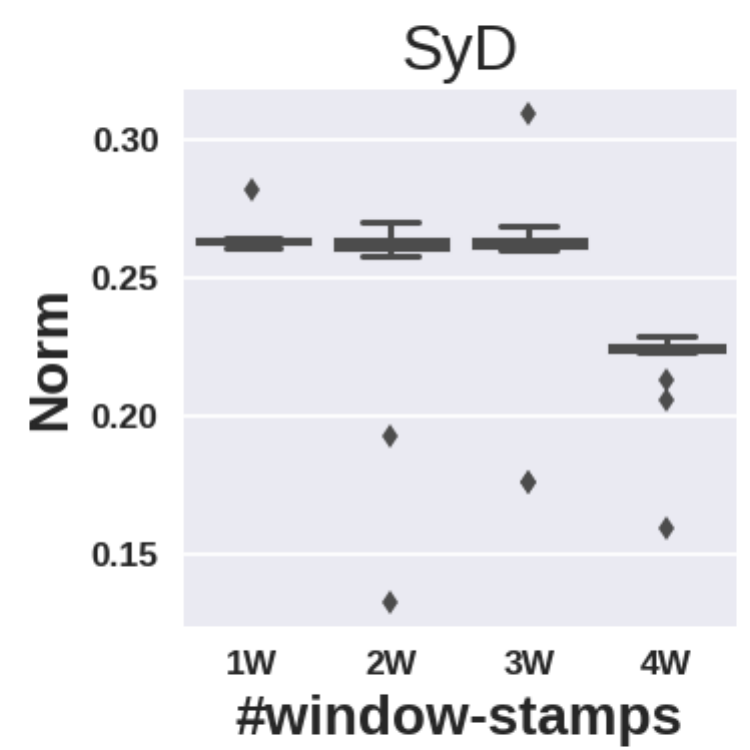}}
    \subfloat{\includegraphics[width=1.56in]{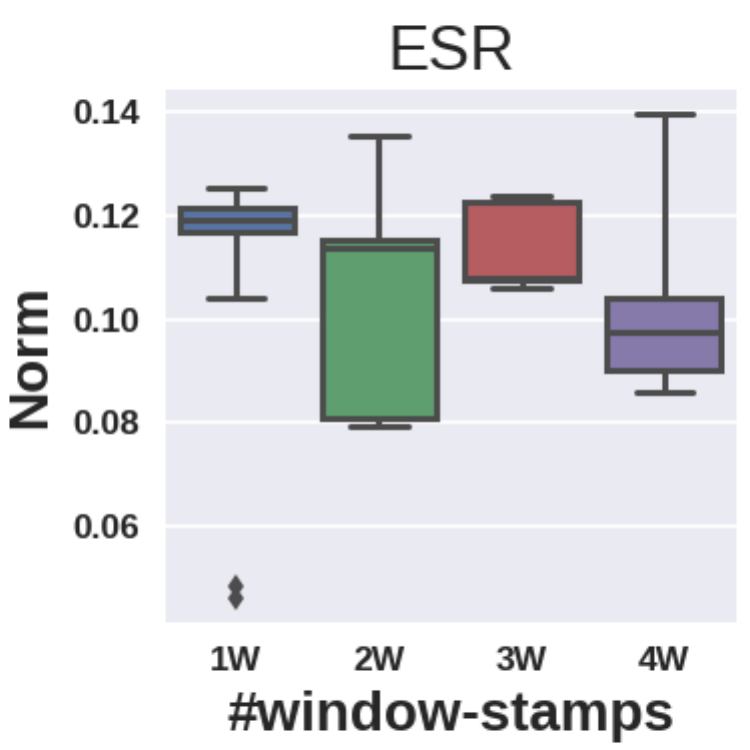}}
    \subfloat{\includegraphics[width=1.56in]{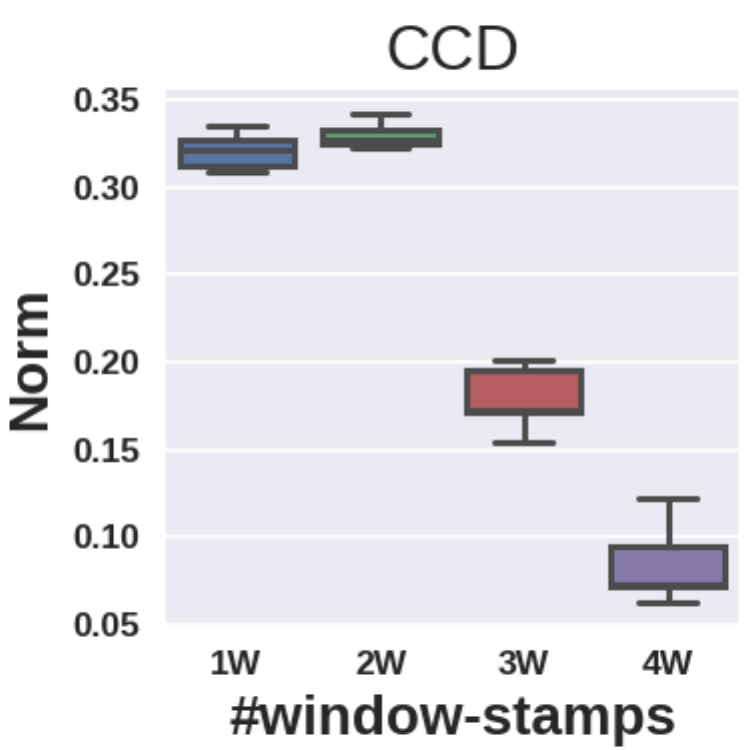}}
    \caption{ Example of error made (the Frobenius Norm) when predicting transition matrice values at 1, 2, 3 and 4 window-stamps ahead in SyD, ESR and CCD.}
    \label{fig:error transition forecasting}
\end{figure}

\subsubsection{Predicting feature vectors \& conditional transitions}
Many regression models can be used to perform the task of predicting the feature vectors. In this paper, we use the $k-$nearest neighbor regression (KNNR) because it is fast and adaptive when series incorporate linear as well as non-linear behavior  \cite{song2017efficient}, \cite{chen2019selecting}. In figures \ref{fig:error feature forecasting} and \ref{fig:error transition forecasting} we show examples of the error made when predicting features and transition matrix values in SyD, ESR and CCD data sets respectively. As depicted in figures \ref{fig:error feature forecasting} and \ref{fig:error transition forecasting}, we can note that, with respect to the Root Mean Square Error (RMSE), the KNNR can predict values with low error rates $1$, $2$, $3$ and $4$ window-stamps ahead. The reported RMSE results show that the KNNR model is able to accurately predict the values at a subsequent time. We note also that, when predicting feature values as well as transition matrices, the error rate does not drastically increase as could have been expected. Instead we note that, when predicting values at further window-stamps, the error quite remains stable if not decreasing. This could be explained by the fact that the feature values (resp. transition matrix values) fluctuate less as time evolves.

\begin{figure}[tbp]
    \centering
    \includegraphics[width=4.65in]{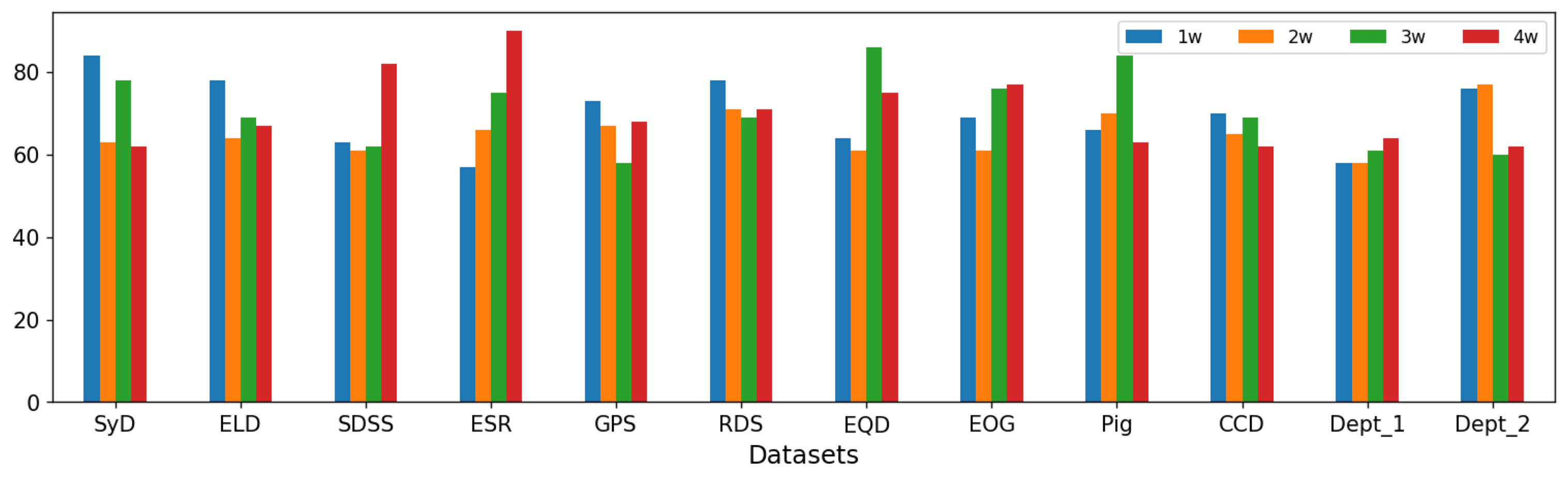} 
    \caption{Results in predicting missing values with respect to the F1 score.}
    \label{fig:missing accuracy}
\end{figure}

\begin{figure}[tbp]
    \centering
    \subfloat{\includegraphics[width=4.7in]{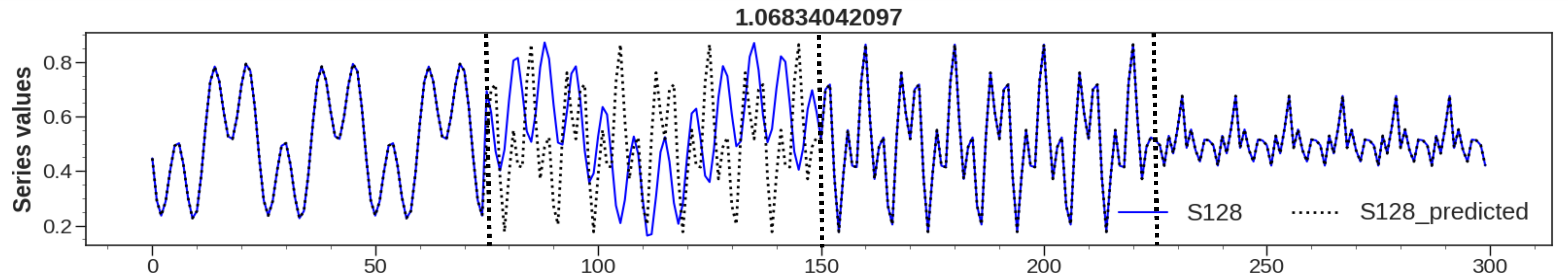}}\\
    \subfloat{\includegraphics[width=4.7in]{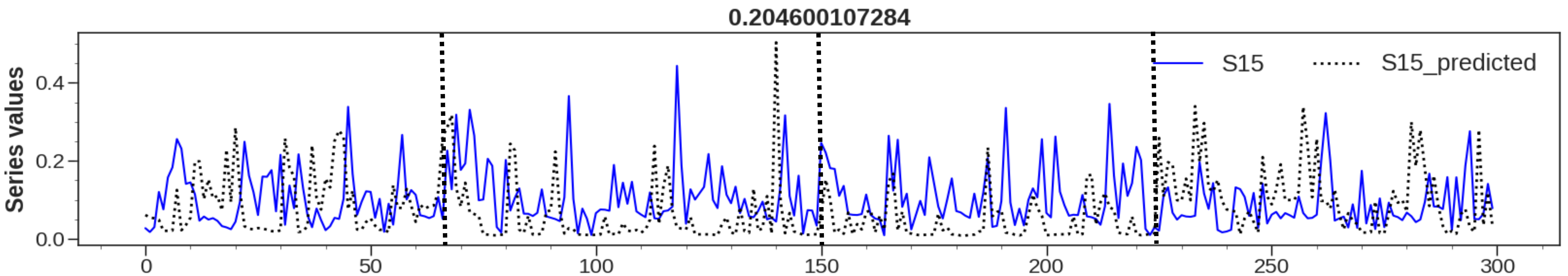}}\\
    \subfloat{\includegraphics[width=4.7in]{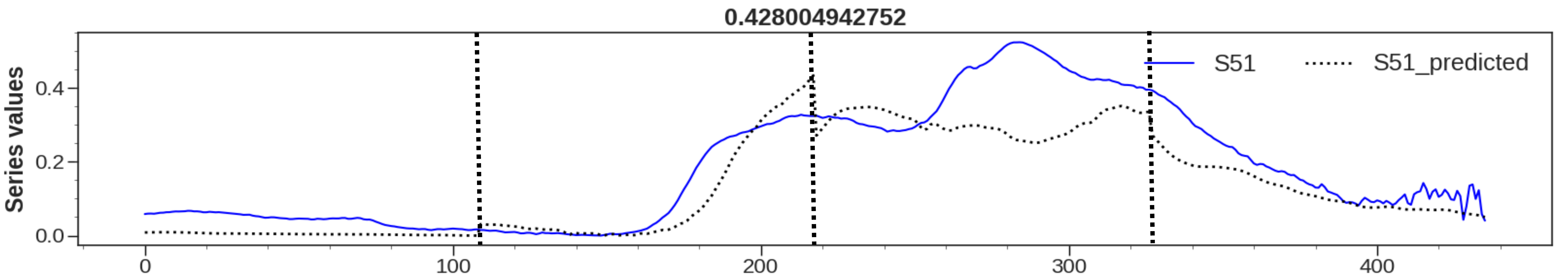}}
    \caption{Example of series prediction. One series was extracted from each of the SyD, SDSS and RDS data sets. The forecast is over the last four window-stamps in each of these data sets. The floating number on each plot corresponds to the Root Mean Square Error. The solid line are the current series values, whereas the dashed lines are the predicted series values.}
    \label{fig:series values forecasting}
\end{figure}

\textbf{Forecasting missing information:} Given a time series $S^i$, to evaluate if it will exhibit missing values, we use the binary codification $0$ and $1$ to refer, respectively to the absence and the presence of a missing value. Hence, for $N$ series for which we predict the presence or absence of missing values, we will have a sequence of $N$ binary values. The obtained predicted sequence is then compared to the real sequence, in terms of whether or not a missing value is exhibited by the $N$ time series. We use the F1 score to evaluate the accuracy of the proposed model in predicting missing values. Fig. \ref{fig:missing accuracy} presents the F1 scores of the proposed model in predicting missing values $1$, $2$, $3$ and $4$ window-stamps ahead in each of the data sets. The reported results show that the model can predict missing values with an average F1 score of $76\%$ in general. 

\textbf{Forecasting series values:} For the purpose of illustration, we randomly picked three time series from the SyD, SDSS and RDS  data sets and present how the model forecasts the series values at (last) four window-stamps.

From Fig. \ref{fig:series values forecasting}, it can be seen how the model enables predicting series's behavior over time. In the first sub-figure (corresponding to SyD) and in the third sub-figure (corresponding to RDS), we can notice that the predicted patterns fit well with the true time series, while in the second (corresponding to SDSS) the predicted patterns correspond to a delayed version of the true time series. Tracking the regime over time clearly helped in predicting series values at subsequent times, although because of the mentioned delays, the predicted values at some time intervals may deviate from the true series values. In our future work, we will further investigate how the identified regimes corresponding to those exhibited within the time series help predict the trend and how to make use of the predicted regimes in real applications such as stock market data analysis.  

\begin{table}[tbp]
    \centering
    \scriptsize{
    \caption{Accuracy of the proposed approach against state-of-the-art methods evaluated by the Mean Absolute Percentage Error.}\label{fig:whole forecasting}
    \begin{tabular}{l|c|c|c|c|c|c|c|c|c|c|c|c|}
         & \multicolumn{4}{c|}{SyD} & \multicolumn{4}{c|}{ELD} & \multicolumn{4}{c|}{SDSS} \tabularnewline \hline
    Model & 1w & 2w & 3w & 4w & 1w & 2w & 3w & 4w & 1w & 2w & 3w & 4w \tabularnewline \hline
    Proposed &\colorbox{gray}{ $\bm{.15}$ }& \colorbox{gray}{ $\bm{.18}$} & \colorbox{gray}{ $\bm{.21}$} & \colorbox{gray}{ $\bm{.28}$} & $.19$ & $.31$ & \colorbox{gray}{ $\bm{.35}$} & \colorbox{gray}{ $\bm{.33}$} & $.28$ & \colorbox{gray}{ $\bm{.28}$} & \colorbox{gray}{ $\bm{.32}$} & \colorbox{gray}{ $\bm{.35}$} \\
    KNNR & $3.02$ & $4.85$ & $9.57$ & $11.7$ & $.33$ & $.44$ & $.47$ & $.62$ & $.82$ & $.86$ & $1.01$ & $.98$ \\
    OrbitMap & $.51$ & $.66$ & $.66$ & $.72$ & $.37$ & $.39$ & $.42$ & $.57$ & $.36$ & $.39$ & $.42$ & $.48$\\
    CNN & $.84$ & $.84$ & $.91$ & $.97$ & \colorbox{gray}{ $\bm{.11}$} & \colorbox{gray}{ $\bm{.21}$} & $.37$ & $.39$ & \colorbox{gray}{ $\bm{.27}$} & $.30$ & $.37$ & $.42$ \\
    VAR & $2.67$ & $3.21$ & $3.41$ & $4.01$ & $.34$ & $.39$ & $.61$ & $.68$ & $1.71$ & $1.71$ & $2.58$ & $2.87$ \\
    NEF & $.52$ & $.59$ & $.78$ & $.84$ & $.48$ & $.48$ & $.62$ & $.61$ & $.28$ & $.31$ & $.36$ & $.44$ \\
    SVR & $2.55$ & $3.27$ & $3.27$ & $4.61$ & $.32$ & $.37$ & $.44$ & $.59$ & $1.21$ & $1.84$ & $2.11$ & $2.54$  \tabularnewline \hline
    & \multicolumn{4}{c|}{ESR} & \multicolumn{4}{c|}{GPS} & \multicolumn{4}{c|}{RDS} \tabularnewline \hline
    Proposed & \colorbox{gray}{ $.17$} & \colorbox{gray}{ $.21$} & $.35$ & \colorbox{gray}{ $.36$} & $.34$ & $.39$ & $.41$ & \colorbox{gray}{$\bm{.44}$} & \colorbox{gray}{$\bm{ .31}$} & $.36$ & \colorbox{gray}{$\bm{.37}$} & \colorbox{gray}{ $\bm{.42}$ } \\
    KNNR & $.23$ & $.38$ & $.57$ & $.63$ & $.63$ & $.67$ & $.84$ & $.88$ & \colorbox{gray}{$\bm{.31}$} & \colorbox{gray}{$\bm{.34}$} & $.39$ & $.53$ \\
    OrbitMap & $.22$ & $.25$ & \colorbox{gray}{ $\bm{.31}$} & $.37$ & \colorbox{gray}{ $\bm{.26}$ }& \colorbox{gray}{ $\bm{.31}$} & $.45$ & $.57$ & $.42$ & $.42$ & $.46$ & $.57$ \\
    CNN & $.41$ & $.47$ & $.54$ & $.54$ & 1$.41$ & $.48$ & $.54$ & $.62$ & $.76$ & $.73$ & $.69$ & $.78$ \\
    VAR & $.20$ & $.27$ & $.39$ & $.61$ & $.52$ & $.63$ & $.67$ & $.74$ & $.32$ & $.38$ & $.44$ & $.51$  \\
    NEF & $.37$ & $.42$ & $.45$ & $.45$ & $.33$ & $.33$ & \colorbox{gray}{ $\bm{.37}$} & \colorbox{gray}{$\bm{.44}$ }& $.36$ & $.39$ & $.57$ & $.63$ \\
    SVR & $.21$ & $.26$ & $.42$ & $.55$ & $.52$ & $.57$ & $.62$ & $.69$ & $.32$ & $.37$ & $.43$ & $.59$ \tabularnewline \hline
    & \multicolumn{4}{c|}{EQD} & \multicolumn{4}{c|}{EOG} & \multicolumn{4}{c|}{Pig} \tabularnewline \hline
    Proposed &  $.19$ & \colorbox{gray}{ $\bm{.22}$ }& \colorbox{gray}{ $\bm{.37}$ }& \colorbox{gray}{ $\bm{.52}$} & \colorbox{gray}{ $\bm{.26}$} & $.35$ & $.37$ & \colorbox{gray}{ $\bm{.37}$} & $.28$ & $.29$ & $.29$ & \colorbox{gray}{ $\bm{.31}$}  \\
    KNNR & $.45$ & $.52$ & $.69$ & $.78$ & $.67$ & $.69$ & $.75$ & $.82$ & \colorbox{gray}{ $\bm{.08}$} & $.19$ & \colorbox{gray}{ $\bm{.26}$ }& $.35$ \\
    OrbitMap & \colorbox{gray}{ $\bm{.16}$ }& $.25$ & $.41$ & \colorbox{gray}{ $\bm{.52}$ }& $.33$ & $.39$ & $.39$ & $.41$ & $.31$ & $.31$ & $.37$ & $.42$ \\
    CNN & $.28$ & $.32$ & $.41$ & $.63$ & $.35$ & \colorbox{gray}{ $\bm{.31}$ }& \colorbox{gray}{ $\bm{.31}$ }& $.46$ & $3.72$ & $3.72$ & $4.25$ & $4.27$ \\
    VAR & $.53$ & $.58$ & $.58$ & $.69$ & $.72$ & $.82$ & $.87$ & $.91$ & $.11$ & $.37$ & $.42$ & $.49$ \\
    NEF & $.27$ & $.32$ & \colorbox{gray}{ $\bm{.37}$} & $.62$ & $.33$ & $.38$ & $.59$ & $.67$ & $.33$ & $.37$ & $.38$ & $.51$ \\
    SVR & $.37$ & $.44$ & $.58$ & $.62$ & $.59$ & $.67$ & $.69$ & $.72$ & \colorbox{gray}{ $\bm{.08}$} & \colorbox{gray}{ $\bm{.14}$ }& $.37$ & $.39$ \tabularnewline \hline
    & \multicolumn{4}{c|}{CCD} & \multicolumn{4}{c|}{Dept\_1} & \multicolumn{4}{c|}{Dept\_2} \tabularnewline \hline
    Proposed & \colorbox{gray}{ $\bm{.24}$} & \colorbox{gray}{ $.27$} & \colorbox{gray}{ $\bm{.37}$} & \colorbox{gray}{ $\bm{.41}$} & \colorbox{gray}{ $\bm{.29}$} & \colorbox{gray}{ $\bm{.37}$ }& \colorbox{gray}{ $\bm{.41}$ }& \colorbox{gray}{ $\bm{.43}$ }& $.29$ & $.33$ & \colorbox{gray}{ $\bm{.35}$} & \colorbox{gray}{ $\bm{.35}$ }\\
    KNNR & $.61$ & $.67$ & $.71$ & $.75$ & $1.71$ & $2.12$ & $2.09$ & $2.17$ & $.35$ & $.39$ & $.41$ & $.46$ \\
    OrbitMap & \colorbox{gray}{ $\bm{.24}$ }& $.29$ & $.41$ & $.52$ & $.42$ & $.51$ & $.51$ & $.57$ & $.32$ & $.39$ & $.41$ & $.47$ \\
    CNN & $.84$ & $.81$ & $.72$ & $.70$ & $.73$ & $.70$ & $.69$ & $.71$ & $2.11$ & $2.31$ & $2.24$ & $2.27$ \\
    VAR & $.78$ & $.78$ & $.82$ & $.84$ & $1.53$ & $1.62$ & $1.62$ & $1.71$ & $.25$ & $.29$ & $.38$ & $.52$ \\
    NEF & \colorbox{gray}{ $\bm{.24}$ }& $.31$ & $.39$ & $.45$ & $.36$ & $.53$ & $.59$ & $.67$ & $.30$ & $.38$ & $.47$ & $.54$ \\
    SVR & $.52$ & $.56$ & $.62$ & $.71$ & $1.41$ & $1.44$ & $1.45$ & $1.42$ & \colorbox{gray}{ $\bm{.24}$ }& \colorbox{gray}{ $\bm{.27}$} & \colorbox{gray}{ $\bm{.35}$ }& $.42$ \tabularnewline \hline
    \end{tabular}
    }
\end{table}

\subsubsection{Benchmark comparison}
Despite its limitation, we conducted a conventional comparative analysis of our proposed model against well-known methods based on error in forecasting series values. The comparative studies include the aggregated time series forecasting proposed in \cite{alzate2013improved}, which makes use of a network-based clustering to group series that are similar and then applies an LSTM over the extracted representative pattern to predict the series values. We chose a network-based clustering because it will permit us to compare our approach (which requires a network clustering as well) to one that does not use any features for forecasting series values. We will use the term NEF, for network-based forecasting, to denote this approach. We also compared the model to other methods, including a k-nearest neighbor regression (KNNR), Support Vector Regression (SVR) 
,  OrBitMap \cite{matsubara2019dynamic}, the Convolutional Neural Network (CNN) and the Vector AutoregRession (VAR). It is important to note that, for the LSTM, we set the number of input neurons to the size of each window. In addition to the input layer, two other layers, with $60$ and $15$ neurons respectively, were considered. For the CNN model, we tune the hyper-parameters by grid search. 

To compare the above models with the proposed approach, we worked only with series for which we have no missing values. Table \ref{fig:whole forecasting} depicts the average Mean Absolute Percentage Error (MAPE) in each of the data sets when forecasting series values one, two, three, and four window-stamps ahead. The MAPEs shown in Table \ref{fig:whole forecasting} illustrate that the proposed model consistently outperforms the other models for long term prediction, that is, in third and fourth window stamps, although the performance of all the methods deteriorates when forecasting several time-stamps ahead. On the other hand, given that the KNNR and the CNN are integrated into our model, the performance difference between the proposed model and these models run alone illustrates the advantage of searching for regime information that explains the changing behaviors observed within the series. These behaviors exhibited by series are hard to capture when looking at the performances of the VAR model, which does not incorporate the changing behavior series may exhibit at different times. 

\section{Conclusion}\label{article2-conclusion}

In this paper, we devised a principled approach for mining and modeling regime shifts in an ecosystem comprising multiple time series. The approach enables us to forecast the series values of this ecosystem at subsequent times. A notable feature of the proposed approach is that it can handle multiple time series dominated by a hidden endless regime shift mechanism. This is accomplished by devising a mapping grid, from which Cox regression and time-dependent transition matrices are utilized to determine regime shift probabilities. The proposed approach is able to automatically uncover various hidden regimes without any prior knowledge about the series under investigation, thanks to the sliding window mechanism we have proposed, which allows the automatic identification of the optimal window size associated with regimes. 

While the proposed approach achieves good results, some aspects still require further work in order to improve the model. For instance,
in our work we assume that the co-evolving time series are dominated by a finite number of regimes having the same length. In some cases, however, the number of regimes and their respective lengths could vary. Accordingly, one future direction is to study in what way our approach can be further expanded to tackle regimes of various lengths in co-evolving time series.


\bibliographystyle{unsrt}

\bibliography{references}

\begin{thebibliography}{10}

\bibitem{sanquer2013smooth}
Matthieu Sanquer~et al.
\newblock A smooth transition model for multiple-regime time series.
\newblock {\em IEEE Trans. on Signal Processing}, 61(7):1835--1847, 2013.

\bibitem{durichen2015multitask}
Robert D{\"u}richen~et al.
\newblock Multitask gaussian processes for multivariate physiological
  time-series analysis.
\newblock {\em IEEE Trans. on Biomedical Engineering}, 62(1):314--322, 2015.

\bibitem{tajeuna2018network}
Etienne~Gael Tajeuna, Mohamed Bouguessa, and Shengrui Wang.
\newblock A network-based approach to enhance electricity load forecasting.
\newblock In {\em 2018 IEEE International Conference on Data Mining Workshops
  (ICDMW)}, pages 266--275. IEEE, 2018.

\bibitem{tan2010day}
Zhongfu Tan, Jinliang Zhang, Jianhui Wang, and Jun Xu.
\newblock Day-ahead electricity price forecasting using wavelet transform
  combined with arima and garch models.
\newblock {\em Applied Energy}, 87(11):3606--3610, 2010.

\bibitem{cirstea2018correlated}
Razvan-Gabriel Cirstea~et al.
\newblock Correlated time series forecasting using multi-task deep neural
  networks.
\newblock In {\em Proc. of the International Conference on Information and
  Knowledge Management}, pages 1527--1530. ACM, 2018.

\bibitem{zhao2018forecasting}
Yi~Zhao~et al.
\newblock Forecasting wavelet transformed time series with attentive neural
  networks.
\newblock In {\em Proc. of the International Conference on Data Mining}, pages
  1452--1457. IEEE, 2018.

\bibitem{matsubara2016regime}
Yasuko Matsubara~et al.
\newblock Regime shifts in streams: Real-time forecasting of co-evolving time
  sequences.
\newblock In {\em Proc. of the International Conference on Knowledge Discovery
  and Data Mining}, pages 1045--1054. ACM, 2016.

\bibitem{matsubara2016non}
Yasuko Matsubara~et al.
\newblock Non-linear mining of competing local activities.
\newblock In {\em Proc. of the International Conference on World Wide Web},
  pages 737--747. International World Wide Web Conferences Steering Committee,
  2016.

\bibitem{wilson2013gaussian}
Andrew Wilson~et al.
\newblock Gaussian process kernels for pattern discovery and extrapolation.
\newblock In {\em Proc. of the International Conference on Machine Learning},
  pages 1067--1075, 2013.

\bibitem{drijfhout2015catalogue}
Sybren Drijfhout~et al.
\newblock Catalogue of abrupt shifts in intergovernmental panel on climate
  change climate models.
\newblock {\em Proc. of the National Academy of Sciences},
  112(43):E5777--E5786, 2015.

\bibitem{chen2012state}
Ming-Hsiang Chen.
\newblock State dependence in the influence of monetary policy regime shifts on
  hospitality index returns.
\newblock {\em International Journal of Hospitality Management},
  31(4):1203--1212, 2012.

\bibitem{dijk2002smooth}
Dick~van Dijk~et al.
\newblock Smooth transition autoregressive models—a survey of recent
  developments.
\newblock {\em Econometric reviews}, 21(1):1--47, 2002.

\bibitem{matsubara2019dynamic}
Yasuko Matsubara~et al.
\newblock Dynamic modeling and forecasting of time-evolving data streams.
\newblock In {\em Proc. of the International Conference on Knowledge Discovery
  and Data Mining}, pages 458--468. ACM, 2019.

\bibitem{cai2015fast}
Yongjie Cai, Hanghang Tong, Wei Fan, and Ping Ji.
\newblock Fast mining of a network of coevolving time series.
\newblock In {\em Proceedings of the 2015 SIAM International Conference on Data
  Mining}, pages 298--306. SIAM, 2015.

\bibitem{deng2021pulse}
Jinliang Deng, Xiusi Chen, Zipei Fan, Renhe Jiang, Xuan Song, and Ivor~W Tsang.
\newblock The pulse of urban transport: Exploring the co-evolving pattern for
  spatio-temporal forecasting.
\newblock {\em ACM Transactions on Knowledge Discovery from Data (TKDD)},
  15(6):1--25, 2021.

\bibitem{chen2018neucast}
Pudi Chen, Shenghua Liu, Chuan Shi, Bryan Hooi, Bai Wang, and Xueqi Cheng.
\newblock Neucast: Seasonal neural forecast of power grid time series.
\newblock In {\em IJCAI}, pages 3315--3321, 2018.

\bibitem{akccay2017short}
H{\"u}seyin Ak{\c{c}}ay and Tansu Filik.
\newblock Short-term wind speed forecasting by spectral analysis from long-term
  observations with missing values.
\newblock {\em Applied energy}, 191:653--662, 2017.

\bibitem{bokde2018novel}
Neeraj Bokde, Marcus~W Beck, Francisco~Mart{\'\i}nez {\'A}lvarez, and Kishore
  Kulat.
\newblock A novel imputation methodology for time series based on pattern
  sequence forecasting.
\newblock {\em Pattern recognition letters}, 116:88--96, 2018.

\bibitem{lubrano2000bayesian}
Michel Lubrano.
\newblock Bayesian analysis of nonlinear time series models with a threshold.
\newblock In {\em Proc. of the International Symposium in Economic Theory},
  volume~11, page~79. Cambridge University Press, 2000.

\bibitem{hochstein2014switching}
Axel Hochstein~et al.
\newblock Switching vector autoregressive models with higher-order regime
  dynamics application to prognostics and health management.
\newblock In {\em Proc. of the International conference on Prognostics and
  Health Management}, pages 1--10. IEEE, 2014.

\bibitem{lutkepohl2005new}
Helmut L{\"u}tkepohl.
\newblock {\em New introduction to multiple time series analysis}.
\newblock Springer Science \& Business Media, 2005.

\bibitem{hallac2017network}
David Hallac~et al.
\newblock Network inference via the time-varying graphical lasso.
\newblock In {\em Proc. of the International Conference on Knowledge Discovery
  and Data Mining}, pages 205--213. ACM, 2017.

\bibitem{li2009dynammo}
Lei Li, James McCann, Nancy~S Pollard, and Christos Faloutsos.
\newblock Dynammo: Mining and summarization of coevolving sequences with
  missing values.
\newblock In {\em Proceedings of the 15th ACM SIGKDD international conference
  on Knowledge discovery and data mining}, pages 507--516, 2009.

\bibitem{che2018recurrent}
Zhengping Che, Sanjay Purushotham, Kyunghyun Cho, David Sontag, and Yan Liu.
\newblock Recurrent neural networks for multivariate time series with missing
  values.
\newblock {\em Scientific reports}, 8(1):1--12, 2018.

\bibitem{habiba2020neural}
Mansura Habiba and Barak~A Pearlmutter.
\newblock Neural odes for informative missingess in multivariate time series.
\newblock In {\em 2020 31st Irish Signals and Systems Conference (ISSC)}, pages
  1--6. IEEE, 2020.

\bibitem{mikalsen2021time}
Karl~{\O}yvind Mikalsen, Cristina Soguero-Ruiz, Filippo~Maria Bianchi, Arthur
  Revhaug, and Robert Jenssen.
\newblock Time series cluster kernels to exploit informative missingness and
  incomplete label information.
\newblock {\em Pattern Recognition}, 115:107896, 2021.

\bibitem{li2016multi}
Yan Li, Jie Wang, Jieping Ye, and Chandan~K Reddy.
\newblock A multi-task learning formulation for survival analysis.
\newblock In {\em Proceedings of the 22nd ACM SIGKDD International Conference
  on Knowledge Discovery and Data Mining}, pages 1715--1724, 2016.

\bibitem{li2017prospecting}
Huayu Li, Yong Ge, Hengshu Zhu, Hui Xiong, and Hongke Zhao.
\newblock Prospecting the career development of talents: A survival analysis
  perspective.
\newblock In {\em Proceedings of the 23rd ACM SIGKDD International Conference
  on Knowledge Discovery and Data Mining}, pages 917--925, 2017.

\bibitem{zhang2019time}
Jianfei Zhang, Shengrui Wang, Lifei Chen, Gongde Guo, Rongbo Chen, and Alain
  Vanasse.
\newblock Time-dependent survival neural network for remaining useful life
  prediction.
\newblock In {\em Pacific-Asia Conference on Knowledge Discovery and Data
  Mining}, pages 441--452. Springer, 2019.

\bibitem{wang2019attributed}
Chun Wang, Shirui Pan, Ruiqi Hu, Guodong Long, Jing Jiang, and Chengqi Zhang.
\newblock Attributed graph clustering: A deep attentional embedding approach.
\newblock {\em arXiv preprint arXiv:1906.06532}, 2019.

\bibitem{lin2007experiencing}
Jessica Lin, Eamonn Keogh, Li~Wei, and Stefano Lonardi.
\newblock Experiencing sax: a novel symbolic representation of time series.
\newblock {\em Data Mining and knowledge discovery}, 15(2):107--144, 2007.

\bibitem{senin2018grammarviz}
Pavel Senin, Jessica Lin, Xing Wang, Tim Oates, Sunil Gandhi, Arnold~P
  Boedihardjo, Crystal Chen, and Susan Frankenstein.
\newblock Grammarviz 3.0: Interactive discovery of variable-length time series
  patterns.
\newblock {\em ACM Transactions on Knowledge Discovery from Data (TKDD)},
  12(1):1--28, 2018.

\bibitem{boutemine2017mining}
Oualid Boutemine and Mohamed Bouguessa.
\newblock Mining community structures in multidimensional networks.
\newblock {\em ACM Transactions on Knowledge Discovery from Data (TKDD)},
  11(4):1--36, 2017.

\bibitem{rosvall2008maps}
Martin Rosvall~et al.
\newblock Maps of random walks on complex networks reveal community structure.
\newblock {\em Proc. of the National Academy of Sciences}, 105(4):1118--1123,
  2008.

\bibitem{lancichinetti2009community}
Andrea Lancichinetti and Santo Fortunato.
\newblock Community detection algorithms: a comparative analysis.
\newblock {\em Physical review E}, 80(5):056117, 2009.

\bibitem{agrawal1998automatic}
Rakesh Agrawal, Johannes Gehrke, Dimitrios Gunopulos, and Prabhakar Raghavan.
\newblock Automatic subspace clustering of high dimensional data for data
  mining applications.
\newblock In {\em Proceedings of the 1998 ACM SIGMOD international conference
  on Management of data}, pages 94--105, 1998.

\bibitem{bouguessa2008mining}
Mohamed Bouguessa and Shengrui Wang.
\newblock Mining projected clusters in high-dimensional spaces.
\newblock {\em IEEE Transactions on Knowledge and Data Engineering},
  21(4):507--522, 2008.

\bibitem{tu2017transnet}
Cunchao Tu, Zhengyan Zhang, Zhiyuan Liu, and Maosong Sun.
\newblock Transnet: Translation-based network representation learning for
  social relation extraction.
\newblock In {\em IJCAI}, pages 2864--2870, 2017.

\bibitem{hamilton2017representation}
William~L Hamilton, Rex Ying, and Jure Leskovec.
\newblock Representation learning on graphs: Methods and applications.
\newblock {\em arXiv preprint arXiv:1709.05584}, 2017.

\bibitem{aalen2008survival}
Odd Aalen~et al.
\newblock {\em Survival and event history analysis: a process point of view}.
\newblock Springer Science \& Business Media, 2008.

\bibitem{yang2016comparative}
Zhao Yang, Ren{\'e} Algesheimer, and Claudio~J Tessone.
\newblock A comparative analysis of community detection algorithms on
  artificial networks.
\newblock {\em Scientific reports}, 6:30750, 2016.

\bibitem{song2017efficient}
Yunsheng Song, Jiye Liang, Jing Lu, and Xingwang Zhao.
\newblock An efficient instance selection algorithm for k nearest neighbor
  regression.
\newblock {\em Neurocomputing}, 251:26--34, 2017.

\bibitem{chen2019selecting}
Ruidi Chen and Ioannis Paschalidis.
\newblock Selecting optimal decisions via distributionally robust
  nearest-neighbor regression.
\newblock In {\em Advances in Neural Information Processing Systems}, pages
  748--758, 2019.

\bibitem{alzate2013improved}
Carlos Alzate and Mathieu Sinn.
\newblock Improved electricity load forecasting via kernel spectral clustering
  of smart meters.
\newblock In {\em 2013 IEEE 13th International Conference on Data Mining},
  pages 943--948. IEEE, 2013.

\end{thebibliography}

\end{document}